%% file: main.tex
\documentclass[twoside]{article}

\usepackage[accepted]{aistats2021}
\usepackage[round]{natbib}

\usepackage[utf8]{inputenc} 
\usepackage[T1]{fontenc}    
\usepackage{hyperref}       
\usepackage{url}            
\usepackage{booktabs}       
\usepackage{amsfonts}       
\usepackage{nicefrac}       
\usepackage{microtype}      

\usepackage{graphicx}
\usepackage{subfigure}
\usepackage{amsmath}
\usepackage{amssymb}
\usepackage{mathtools}
\usepackage{wrapfig}

\newtheorem{theorem}{Theorem}

\newtheorem{lemma}{Lemma}

\usepackage{algorithm}
\usepackage{algpseudocode}

\DeclareMathOperator*{\argmax}{arg\,max}
\DeclareMathOperator*{\argmin}{arg\,min}
\def\N{\mathcal{N}}
\def\E{\mathbb{E}} 
\def\P{\mathbb{P}}
\def\R{\mathbb{R}}
\def\KG{\text{KG}}
\def\yn1{y^{n+1}}
\def\mun1{\mu^{n+1}}
\def\n1{^{n+1}}
\def\tXn{\tilde{X}^n}

\def\CONBO{{\text{\CONBOt}}}

\def\MI{\text{MI}}
\def\ES{\text{ES}}
\def\PES{\text{PES}}
\def\MES{\text{MES}}

\def\ConBO{\text{ConBO}}
\def\KG{\text{KG}}
\def\yn1{{y^{n+1}}}
\def\mun1{\mu^{n+1}}
\def\n1{^{n+1}}
\def\tXn{\tilde{X}^n}
\def\txn1{\tilde{x}\n1}
\def\CONBO{{\text{ConBO}}}
\def\MI{\text{MI}}
\def\ES{\text{ES}}
\def\PES{\text{PES}}
\def\MES{\text{MES}}
\def\xs{{x^*}}
\def\xsi{{x^*_{s_i}}}
\def\ysi{{y^*_{s_i}}}
\def\VoI{\text{VoI}}
\def\sigt{\tilde{\sigma}}

\def\Zt{\tilde{Z}}
\def\mul{\underline{\mu}}
\def\sigl{\underline{\sigma}}

\def\qed{\hfill $\square$}

\DeclarePairedDelimiter{\ceil}{\lceil}{\rceil}
\begin{document}

%

%

\twocolumn[

\aistatstitle{Practical Bayesian Optimization of Objectives with Conditioning Variables}

\aistatsauthor{ Michael Pearce \And Janis Klaise \And  Matthew Groves}

\aistatsaddress{ Centre for Complexity Science\\
  Warwick University\\
  Coventry, UK \\ \And
      Seldon Technologies\\
      London, UK \\ \And
    Centre for Doctoral Training in\\ Mathematics for Real-World Systems \\
  Warwick University\\
  Coventry, UK \\ } ]

\begin{abstract}
 Bayesian optimization is a class of data efficient model based algorithms typically
focused on global optimization. We consider the more general case where a user is
faced with multiple problems that each need to be optimized conditional on a state
variable, for example given a range of cities with different patient distributions,
we optimize the ambulance locations conditioned on patient distribution. Given partitions of CIFAR-10,
we optimize CNN hyperparameters for each partition. Similarity
across objectives boosts optimization of each objective in two ways: in modelling by
data sharing across objectives, and also in acquisition by quantifying how a single
point on one objective can provide benefit to all objectives. For this we propose a framework
for conditional optimization: ConBO. This can be built on top of a range of acquisition functions
and we propose a new Hybrid Knowledge Gradient acquisition function. The resulting method is intuitive and
theoretically grounded, performs either similar to or significantly better than recently published works
on a range of problems, and is easily parallelized to collect a batch of points.
\end{abstract}

\input{secs/0-intro1.tex}

\input{secs/1-probspec.tex}

\input{secs/3-method.tex}

\input{secs/5-Experiments-Neurips}

\input{secs/7-Conclucsion-Neurips}

\section*{Acknowledgements}
We would like to thank Ayman Boustati for helpful discussions and code review. We are grateful for the emotional feline support of Fred, Tonks, Luna, and her highness Ogden Balmer Von Stroppleslouth the Third.
\bibliography{CONBO}

\onecolumn
\appendix

\input{sm_secs/0.5theory}

\input{sm_secs/0-hybrid-KG}

\input{sm_secs/2-Experiment_settings}

\input{sm_secs/4-parallel}


\input{sm_secs/0-GLOBAL-OPTIM}

\input{sm_secs/1-ES_MES}

\end{document}

%% file: secs/0-intro1.tex
\section{Introduction}


Expensive black box functions arise in many fields such as fluid simulations~\citep{picheny2013nonstationary}, engineering wing design~\citep{jeong2005efficient},
and machine learning parameter tuning~\citep{snoek2012practical}. In this work we
consider the more general case where the expensive function must be optimized for
a range of settings, or conditioned on a \emph{state}, often called a context,
nuisance variable or source. This has many applications as follows.


\textbf{Physics simulators:} the optimal packing fraction of particles in a container
varies with particle size~\citep{ginsbourger2014bayesian}.
A nuclear fusion reactor has multiple plasma states each requiring unique controls~\citep{char2019offline, chung2020offline} .

\textbf{Algorithm Parameter Tuning:}
given a set of graphs one may find the best coloring algorithm for each graph~\citep{Smith-Miles2014}.
For a selection of datasets one can optimize model training hyperparameters~\citep{bardenet2013collaborative}  
for each dataset. We consider CNN optimizer parameters for partitions of CIFAR-10.

\textbf{Robust Engineering:}
when designing a wing for a range of environment conditions and flow angles, a safety engineer
needs to find flow angle of \emph{worst case} stresses of the wing for each environment.

\textbf{Logistics:}
given a range of warehouses spread over a country that each face different sales demand, one
must optimize the stock control policy of each warehouse~\citep{poloczek2016warm}.
Given a range of cities with different population distributions, for each
city one must optimize ambulance locations~\citep{dong2017empirically}.
We also consider these two applications.

In this work we shall refer to variables that are 
to be optimized as \emph{actions}, though parameters, decision variables,
and solutions are often used.
There are many variations of the conditional setting. 
In contextual optimization~\citep{paul2018contextual}, at each
iteration a context (or state) is passed
to the optimization algorithm, the algorithm
chooses an action and the reward is returned.
Studied applications include drug design~\citep{krause2011contextual}
%
\begin{figure}
    \centering
    \includegraphics[width=0.45\textwidth]{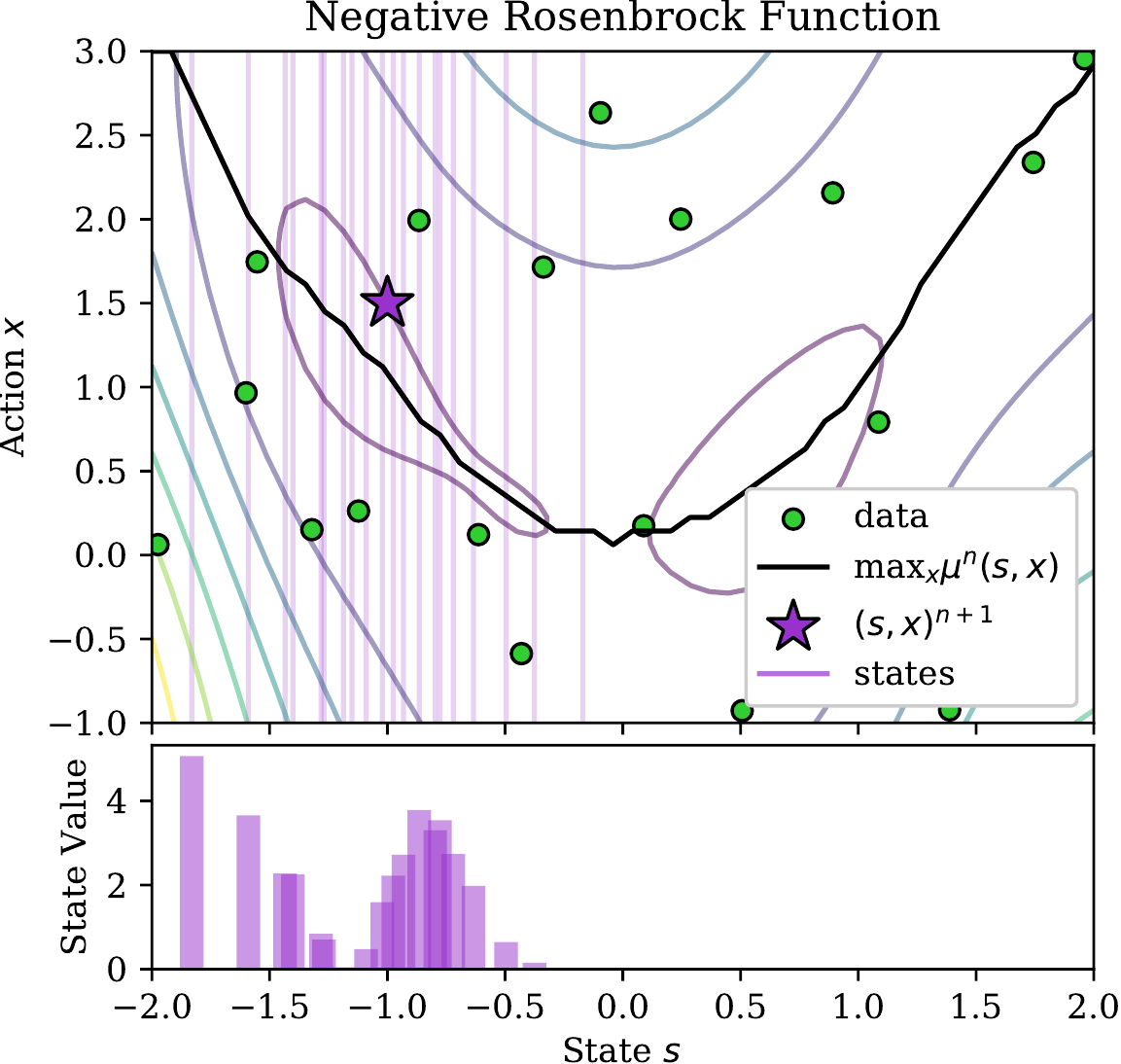}
    \caption{Top: GP model. Each state (vertical slice)
    is an objective function to maximise. At each iteration the algorithm must determine $(s,x)\n1$ that provides
    information about the optima of all states (black line).
    Bottom: the acquisition function value of each
    local Monte-Carlo state using hybrid Knowledge Gradient.}
    \label{fig:KG_per_state}
\end{figure}
%
%
in which an algorithm can visit molecules (states) in a round-robin fashion and chooses
a test chemical (action) at each iteration.
Optimization with warm starts, dynamic objectives, transfer learning~\citep{MoralesBranke15, feurer2015initializing, poloczek2016warm, perrone2018scalable}
may be interpreted as contextual optimization where the state remains
constant for multiple time steps in which multiple actions can be evaluated.
Such methods have been applied to the aforementioned many warehouses
example and to hyperparameters of neural networks for sequentially arriving datasets.

Many applications in these works may also be viewed as serialised versions of
conditional optimization problems. In drug design, an
experimenter may be free to choose both the molecule and
test chemical for each experiment, in hyperparameter tuning,
if a collection of datasets is already in user possession, the user is free to choose
both the dataset and parameters to train next. Both the context/state and the action
is under algorithm control and the goal is to \emph{maximise all states simultaneously}.
See Figure~\ref{fig:KG_per_state}.

In multi information source, or multi-fidelity optimization~\citep{huang2006sequentialMF,poloczek2017multi,kandasamy2017multi} 
one has multiple functions over action space, each one
corresponds to a source (or state) and has a different
query cost. The goal is to optimize an expensive source
(target state) by sequentially evaluating actions on 
cheaper sources while exploiting similarity across sources.
The conditional setting is a generalization where all sources
are targets and (for simplicity) we assume all sources/states
have equal query cost.

Past works have referred to this setting as profile optimization~\citep{Ginsbourger2014}, 
conditional optimization~\citep{sambakhe2019conditional}, offline contextual optimization~\citep{char2019offline}, collaborative tuning~\citep{bardenet2013collaborative}, and
continuous multi-task optimization~\citep{pearce2018continuous}. In this
work we also adopt \textit{conditional optimization}.

This paper makes the following contributions:
\begin{enumerate}
    \item A framework for conditional Bayesian optimization with theoretical guarantees: $\CONBO$
    \item A new competitive global optimization method that is cheap to compute: Hybrid
Knowledge Gradient
    \item Extensive evaluation on fully open source conditional optimization
    problems from various domains.
\end{enumerate}

\section{Related Work}
We focus on conditional algorithms. We define such algorithms
as those that consider a set of functions, each corresponding to a state. For each iteration, the algorithm is free to choose both the state and action. The goal is to find the
peak of each function, all of the conditional optima, see Figure~\ref{fig:KG_per_state}. 

The SCoT algorithm~\citep{bardenet2013collaborative} consider discrete states and visits
states in a round-robin fashion and chooses actions using
expected improvement. The authors consider hyperparameters of machine learning algorithms
for multiple datasets, all of which are considered to be available upfront.
As SCoT predetermines the sequence of states it is unnecessarily
restrictive, and also cannot be applied to problems with continuous states.

The Profile Expected Improvement (PEI) algorithm~\citep{ginsbourger2014bayesian} does consider continuous state
and proposes an acquisition function for the next state action pair by
measuring expected improvement of a new outcome over the best predicted outcome
within the same state. The recent PEQI method~\citep{sambakhe2019conditional} extends PEI
to noisy problems using SKO~\citep{huang2006sequential}. 
These algorithms significantly improve upon SCoT by allowing the algorithm to
\emph{dynamically} determine the state as well as the action of each sampled
state action pair. However, the proposed acquisition functions only
quantify the benefit of a new sample by the effect it may have at
finding the maxima within the sampled state,
it does not account for how a sample may help to find the maxima
of similar states.

The REVI algorithm~\citep{pearce2018continuous} aims to improve upon these methods and 
explicitly quantifies the benefit a new sample will have for finding the maxima
of all states. At each iteration, the method discretizes both state and action
domains and treating a continuous problem as a discrete one. Given a
hypothetical state action point to sample, the Knowledge Gradient by discretization (over actions)
quantifies how the new sample will benefit each state in the discretized set of states.
The sum of these benefits provides an estimate of overall benefit. The method suffers
the curse of dimensionality, increasing dimension leads to exponentially increasing
discrertization size with corresponding exponential space and time requirements.
Using a smaller discretization suffers discretization errors that can largely undermine 
any benefit of accounting for all states and actions.

Most recently, the MTS method~\citep{char2019offline} uses a novel kernel with
a length scale that varies across states. This is combined with a multi-task
Thompson sampling method for collecting new data. Note that this method also 
requires a discretization, over real state-action space or Fourier space,
and has cubic cost with the size of the discretization. Further, Thompson sampling,
like earlier works, does not account for how one sample provides benefit for
optimizing all states, fundamental structure of conditional problems.

While REVI and MTS appear to be the strongest algorithms,
both suffer from practical issues, REVI is exponentially expensive
and MTS does not utilise the structure of the problem.

In this work, we propose a method that both exploits the basic structure
of conditional optimization, like REVI, while also being much more practical and
computationally favourable, like SCoT, PEI or PEQI, and therefore amenable to
much more challenging real world problems.



%% file: secs/1-probspec.tex
\section{Problem Statement}\label{sec:probdef}

Adopting the convenient terminology of bandits, we assume that we have
an expensive black box function that takes as input both a
\emph{state} in state space $s\in S$ and an \emph{action}
in action space $x \in X$ either of which may be continuous
or discrete. The expensive function returns a noisy scalar \emph{reward},
\begin{equation}
f(s,x):S \times X \to \R.
\end{equation}
There is a distribution over states $\P[s]$ that encodes the priority
or weighting of states. Given a budget of $N$
function calls, we sequentially choose points $(s, x)$
and observe a noisy reward $f(s,x)$. The objective is to learn the
best action for every state, or
a policy $\pi:S \to X$, that maximises expected output for all states,
\begin{equation} \label{eq:TrueTotalReward}
\text{Total Reward} = \int_S \E[f(s, \pi(s))] \P[s] ds,
\end{equation}
where the expectation is over the stochasticity in rewards.
If there is only a single state, the reward function is a single objective,
the policy is a single action, and the problem reduces to global optimisation.
If $S$ is a finite set, the integral reduces to a summation. In this work
we consider the following applications:

\textbf{CNN hyperparameters:} \label{sec:probdec_cnn} 
the CIFAR-10 dataset  contains 10 classes, we split this into five datasets, $S=\{1, ...,5\}$,
of two classes each with uniform state density $\P[s]=1/5$.
Using a 5 layer CNN, we optimize dropout rates, batch size and
Adam parameters, $X\subset \R^7$. $f(s,x)$ is the validation accuracy.

\textbf{Ambulances in a square:}~\citep{dong2017empirically}\label{sec:probdef_ambulances}
given a range of 30km$\times$30km city maps, each city is parameterized by the mode of its
population distribution $s\in S=[0, 30]^2$. $\P[s]$ is a truncated
Gaussian around the centre of the map; most cities, like Paris,
are densest in the centre while others are densest at a coast
line like Singapore. The action space is the 2D coordinates
of three ambulance bases, $x\in X=[0, 30]^6$, and for each city $s$
one must find bases $x$ to minimise journey
times $f(s, x)$ to patients over a simulated day.

\textbf{Assemble to order:}~\citep{xie2016bayesian} \label{sec:probdef_ato}
a company owns many stock warehouses and each one
faces a different level of demand $s\in [0.5, 1.5]$.
The distribution of demand $\P[s]$ is uniform.
Stock is depleted as customer orders arrive and are fulfilled,
and stock levels are controlled by setting targets 
for replenishment, $x\in X = [0, 20]^8$. The objective, $f(s,x)$, is sales profit minus holding costs over a month.

%% file: secs/3-method.tex
\section{The Conditional Bayesian Optimization Algorithm}
We first discuss the fitting of the Gaussian process model and the policy.
We then motivate the acquisition value for a single state
and how this is integrated over states yielding the acquisition function.
Integrating over states increases the computational burden hence we propose
Hybrid Knowledge Gradient as a solution.

At a stage after having observed $n$ data points, 
$\{(s^i, x^i, y^i) \}_{i=1}^n$ where $y^i=f(s^i, x^i)$,
we fit a Gaussian process from the joint space $S\times X$ to outputs $y\in \R$.
Let $\tXn = ((s,x)^1,...,(s,x)^n)$ and $Y^n = (y^1,...,y^n)$.
A Gaussian process is defined by a prior mean and prior covariance function, $\mu^0(s,x)$, $k^0((s,x),(s',x'))$
which are chosen for each application,
for more information see \cite{rasmussen2006gaussian}.
After observing $n$ data points, let $K = k^0(\tXn, \tXn)\in \R^{n\times n}$, the posterior mean is given by 
\begin{align}
    \mu^n(s&,x) = \mu^0(s,x) \nonumber\\
       &+k^0(s, x, \tXn)  \big(K + \sigma^2_0I\big)^{-1} (Y^n-\mu^0(\tXn))
       \label{eq:post_mean}
\end{align}
and the posterior covariance is given by
\begin{align}
     k^n(s,x,&s',x') =k^0(s, x, s',x') \\ 
     & + k^0(s, x, \tXn) \big(K + \sigma^2_0I\big)^{-1} k^0(\tXn, s',x').\nonumber
\end{align}
For state $s$, the predicted optimal action $x$ defines a policy
\begin{equation} \label{eq:GP_policy}
    \pi^n(s) = \argmax_x \mu^n(s, x).
\end{equation}
Although we don't use them in our experiments, a common approach
for applying GPs to higher dimensions is additive kernels~\citep{kandasamy2015high, krause2011contextual}. 
These assume the function decomposes into a sum of functions, each one
depending on an exclusive subset of variables e.g. $f((x_1, x_2)) = f_1(x_1) + f_2(x_2)$.
If the state and action variables are in different subsets the policy becomes
$\pi^n(s) = \argmax_x \mu_1^n(s) + \mu^n_2(x) = \argmax_x \mu^n_2(x)$, it is independent
of the state. Additive structure assumes \emph{no interaction} between variables while conditional
optimization is only necessary \emph{with interaction}. Thus in settings
with additive structure, one may fix all non-interacting state and action variables
and apply conditional optimization only to additive components containing both
state and action variables.

Collecting a new data point $y\n1$ at $(s,x)\n1$ will update the Gaussian process model for all points
$S\times X$, updating the predicted peak of all states affecting the policy.
To construct an acquisition function for conditional optimization,
we start by looking for standard acquisition functions that account for how the
model changes at \emph{unsampled} locations $(s',x')\neq (s,x)\n1$.
Specifically, the popular Expected Improvement (EI)~\citep{Jones1998}
and upper confidence bound (UCB)~\citep{kandasamy2016gaussian} methods
are both functions of \emph{the mean and variance at the sampled point only}, $\mu^n((s,x)\n1)$,
$k^n((s,x)\n1, (s,x)\n1)$.
Methods that utilise other points in the domain
include Entropy search (ES and PES) that measures the mutual
information between the new output $\P[y\n1|x\n1]$ and the (induced)
location of the peak $\P[x^*|\tXn, Y^n]$, Max-value entropy search 
(MES)~\citep{wang2017max} measures mutual information
between the new output $\P[y\n1|x\n1]$ and the largest output
$\P[\max y|\tXn, Y^n]$, and Knowledge Gradient (KG)~\citep{Frazier2009}
that measures the expected new peak posterior mean $\E[\max \mu\n1(x)]$ caused
by a new $y\n1$ at $x\n1$. Unlike EI and UCB, we classify ES, MES, KG as ``globally aware''. 


\begin{figure*}[t]
    \centering
    \includegraphics[width=0.95\textwidth]{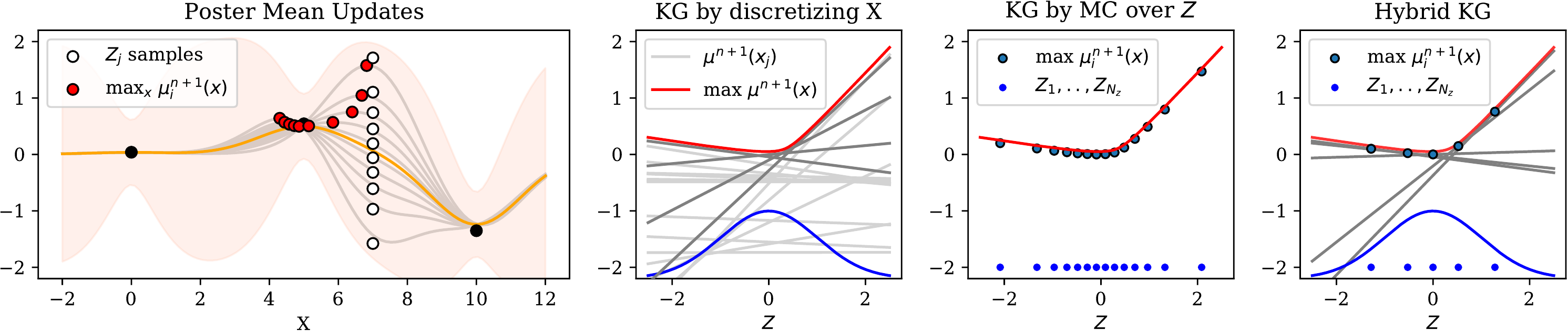}
    \caption{
    \label{fig:KG_demo}
    Methods for computing $\KG(x\n1)$ at $x\n1=7$.
    Left: $\mu^n(x)$ and samples of $\mu\n1(x)$ determined by a scalar $Z\sim N(0,1)$.
    Centre-left: $\KG_d$ replaces $X$ with up to 3000 points $x_i\in X_d$ and $\mu\n1(x_i)$ is linear in $Z$.
    Centre-right: $\KG_{MC}$ samples up to 1000 functions $\mu\n1(x)$ functions and maximises each of them numerically.
    Right: $\KG_h$ samples up to 5 functions $\mu\n1(x)$ and maximizes them numerically, 
    the $\argmax$ points $x^*_1,..,x^*_5$ are used as $X_d$ in $\KG_d$.
     }
\end{figure*}

Each state $s_i$ defines a single global optimization problem over $x\in X$,
using $\ES$, $\PES$, $\MES$ or $\KG$, the acquisition value for state $s_i$
given a sample elsewhere at $(s,x)\n1$ may be computed. In this work we adopt $\KG$ for 
its Bayesian decision theoretic derivation that extends seamlessly to the conditional
setting. For KG, the benefit for state $s_i$ given a sample at $(s, x)\n1$ is the
expected increase in predicted reward for the best action
within state $s_i$. We denote this as $\text{KG}_c(\cdot)$ given by
\begin{align}
\text{KG}_c(s_i; (s,x)\n1) =& \\
\E_{y\n1}[\max_{x'} \mu\n1(s_i,x')&|(s,x)\n1] - \max_{x''}\mu^n(s_i,x''). \nonumber
\end{align}
We discuss numerical evaluation of $\KG_c(\cdot)$ in Section \ref{sec:hybrid_KG}. Similar expressions for entropy methods, 
$\ES_c(\cdot), \, \PES_c(\cdot), \, \MES_c(\cdot)$, are derived in the Supplementary Material.
Integrating over states $s_i$ yields the total acquisition value
\begin{align}
    \int_S\text{KG}_c(s; (s,x)\n1) \P[s]ds.
\end{align}
For discrete $S$ the integral is replaced by summation.
For continuous $S$, the integral over states $s$ cannot be computed analytically and so we use
Monte Carlo with importance sampling. When using a kernel that factorises
$k(s,x, s',x') = \sigma_0^2 k_S(s,s')k_X(x, x')$, like squared exponential
or Mat\'ern, similarity across states is encoded in $k_S(s,s')$.
This naturally leads to the proposal distribution
$q(s|s\n1) \propto k_S(s,s\n1)$.
In our continuous state experiments, we use the Mat\'ern kernel and a Gaussian proposal distribution
with mean $s\n1$ and the state kernel length scales, $l_s$, as standard deviations,
\begin{equation}
    q(s|s\n1) \sim \N(s|s\n1, \text{diag}(l_s^2)).
\end{equation}
We generate $n_s=20$ samples $S_{MC} = \{s_1,...,s_{n_s}\}$, finally the acquisition
function is
\begin{align}
    \CONBO(s\n1,x\n1) =&\\ \sum_{s_i\in S_{MC}} \frac{\P[s_i]}{q(s_i|s\n1)}& \text{KG}_c(s_i; (s,x)\n1). \nonumber 
\end{align}
Figure~\ref{fig:KG_per_state} shows a set of sampled states and the $\KG_c(\cdot)$ for each one.
Each KG term directly measures increase in predicted reward for one state and the above sum measures
reward increases across all states, this is a direct surrogate for maximizing the
true objective (Equation~\ref{eq:TrueTotalReward}). For entropy based methods,
$\CONBO$ becomes a sum of Shannon information units thereby indirectly
optimizing the true objective.
The randomly sampled states $S_{MC}$ may be resampled with each call
to $\CONBO(s, x)$ and gradients estimated enabling the optimal $(s,x)\n1$
to be found with a stochastic gradient ascent optimizer such as Adam~\citep{kingma2014adam}. Selecting each point according to maximising $\CONBO$ is
also myopically optimal in a value of information framework:
\begin{theorem}
Let $(s^*, x^*) = \argmax \CONBO(s, x)$ be a point chosen for sampling. $(s^*, x^*)$ is also the point that
maximises the myopic Value of Information, the increase in predicted policy reward.
\end{theorem}
Further, in finite search space, with an infinite sampling budget all points will be sampled infinitely:
\begin{theorem}
Let $S$ and $X$ be finite sets and $N$ the budget to be sequentially
allocated by $\CONBO$. Let $n(s,x, N)$ be the number of samples
allocated to point $s,x$ within budget $N$. Then for all $(s,x) \in S \times X$
we have that $\lim_{N\to \infty} n(s,x, N)=\infty.$
\end{theorem}
The law of large numbers ensures that the algorithm learns the true
expected reward for all points. Proofs are given in the Supplementary Material.

\subsection{Hybrid Knowledge Gradient}\label{sec:hybrid_KG}
By definition, $\KG$ is more expensive than EI and UCB. Further,
the function $\KG_c(s_i,(s,x)\n1)$ must be computed once for each
sampled state $s_i$, the computational cost is therefore $n_s$ times
the global acquisition function equivalent. To alleviate this 
cost, we propose a novel, efficient algorithm for computing KG.
In the following section we assume constant $s$ for brevity, reducing to the global optimization setting. Given a hypothetical
location $x^{n+1}$, KG quantifies the value of a new hypothetical
observation $y^{n+1}$ by the expected increase in predicted
reward of the optimal action, i.e. the expected increase in the peak of the posterior mean
\begin{equation}\label{eq:KG_global}
    \KG(x\n1) = \E_{\yn1}\big[\max_{x'} \mun1(x')\big| x\n1\big] - \max_{x''} \mu^n(x'').
\end{equation}
However, $\max_{x'} \mun1(x')$ has no direct formula
and approximations are required which we describe next. At time
$n$, the new posterior mean is unknown, however, it may be written as
$\mu^{n+1}(x) = \mu^n(x) + \tilde\sigma(x; x\n1)Z$ where 
$\tilde\sigma(x; x\n1)$ is a deterministic function and the scalar $Z\sim\N(0,1)$
captures the randomness of $y\n1$, see SM. Previously, 
$\KG(x)$ has been computed in the following two ways.


\textbf{KG by discretization}~\citep{Frazier2009, scott2011correlated}:
in Equation~\ref{eq:KG_global}, the maximizations
may be performed over a discrete set of $d$ points $x'\in X_d \subset X$. 
Denoting $\underline{\mu}=\mu^n(X_d)\in\R^d$ and $\underline{\tilde\sigma}(x\n1) = \tilde\sigma(X_d; x\n1)\in\R^d$, then
$$
\KG_d(x\n1) = \E_Z\left[ \max \{\underline \mu + \underline{\tilde\sigma}(x\n1)Z \} \right] - \max \underline{\mu}.
$$
The $\max \{\underline \mu + \underline{\tilde\sigma}(x\n1)Z \}$ is a piece-wise
linear function of $Z$ and the expectation over $Z\sim N(0,1)$ is analytically tractable.
The output is a \emph{lower bound} of the true $\KG(x)$ over the continuous set.
The MISO algorithm~\citep{poloczek2017multi}
used $\KG_d$ with 3000 uniformly random distributed points.
This method suffers the curse of dimensionality, $X_d$ must grow
exponentially with dimension of $X$. Even when using a dense
discretization, $X_d$ may contain many unnecessary points
$x_i$ that do not form part of $\max \mun1(X_d)$,
see Figure~\ref{fig:KG_demo} centre-left plot.


\textbf{KG by Monte Carlo}~\citep{wu2017discretization, toscano2018bayesian}:
given $x\n1$, the method samples up to
$n_z=1000$ values of $Z$. For each $Z_j$, construct $\mun1_j(x) = \mu^n(x) + \tilde\sigma(x; x\n1)Z_j$
and the peak output value of each $\mun1_j(x)$ is found using a continuous
numerical {\tt Optimizer()} (e.g. L-BFGS, CG), 
\begin{align*}
\KG_{MC}(x\n1) &= \\
\frac{1}{n_s}\sum_j &\underset{x'}{\text{\tt{Optimizer}}}\big(\mu^n(x') + \tilde\sigma(x'; x\n1)Z_j\big) \\
&- \underset{x''}{\tt{Optimizer}}\big(\mu^n(x'')\big).
\end{align*}

The resulting average is an \emph{unbiased}
estimate of true $\KG(x)$ and scales better to higher dimensional $X$ as the univariate $Z$
is discretized by Monte Carlo samples instead of $X$.
However, similar $Z_j$ values will have similar $\max\mu^{n+1}_j(x)$ making
many calls to {\tt Optimizer()} redundant. See Figure~\ref{fig:KG_demo} centre right.


We instead propose a simple natural mixture of the above two approaches that
scales to higher dimensional $X$ and reduces redundant {\tt Optimizer()} calls.

\textbf{Hybrid KG:} given $x\n1$, first following $\KG_{MC}$
we use $n_z=5$ values of $Z_j$ and for each $\mu^n(x) + \tilde\sigma(x; x\n1)Z_j$
we find the argmax $x^*_j$ using {\tt Optimizer()}. Second, following $\KG_d$ we treat the set of
peak locations as a \emph{dynamic optimized discretization} $X^*_d = \{x^*_1,...,x^*_{n_z}\}$
to analytically compute a lower bound of the true $\KG(x)$. Let 
$\underline{\mu}^*=\mu^n(X^*_d)\in\R^{n_z}$ and $\underline{\tilde\sigma}^*(x\n1) = \tilde\sigma(X_d^*; x\n1)\in\R^{n_z}$, then
$$
\KG_h(x\n1) = \E_Z[\max\underline{\mu}^{*} + \underline{\tilde\sigma}^*(x\n1)Z\big] - \max\underline{\mu}^{*}.
$$
Compared to $\KG_d$, the hybrid method removes redundant points
in the discretization $X_d$, all $X^*_d$ points contribute to $\max\mu^{n+1}(X^*_d)$
and there are far fewer points.
Compared to $\KG_{MC}$, instead of sampling many $Z_j$ and finding many $x^*_j$, 
far fewer $Z_j$ are sampled and fewer need to be $x^*_j$ found.

To ensure asymptotic convergence, in a discrete domain, we require that the acquisition
function is non-negative, $\KG_h(x)\geq 0$, and the acquisition function
is zero where GP variance is zero, $\KG_h(x)=0 \iff k^n(x,x)=0$. 
Therefore, always choosing $x\n1= \argmax \KG_h(x)$ ensures only 
points with GP variance will be revisited until all points have no
variance i.e. the true function is known for all points.
We can ensure these properties by setting $n_z\geq2$ and at least one $Z_j$
is equal to zero.
\begin{theorem}
Let $n_z\geq 2$ and let $\underline{Z} = \{Z_j|j=1,...,n_z\}$. If $0 \in \underline{Z}$
then $KG_h(x)\geq 0$ for all $x\in X$ and if $x$ is sampled infinitely often $\KG_h(x) = 0$.
\end{theorem}
Proof is in the SM.
The $Z_j$ values can be fixed, for $n_z=5$ we use equal Gaussian quantiles
$\underline{Z}=\big\{\Phi^{-1}(0.1), \Phi^{-1}(0.3),\dots, \Phi^{-1}(0.9)\big\}$ where
$\Phi(\cdot)$ is the Gaussian CDF.
Using quantile spacing and odd $n_z$ ensures $Z_j=\Phi^{-1}(0.5)=0$ is
included which satisfies the assumptions of asymptotic convergence.
See Fig.\ref{fig:KG_demo} (right).

The computational complexity of a single call to $\CONBO$
requires the posterior variance ($O(n^2)$) and $n_sn_z$
runs of {\tt Optimizer()}. Assuming {\tt Optimizer()} internally
makes $n_{calls}$ to the posterior mean $O(n)$, $\CONBO$ total complexity is
$O(n^2 + nn_{calls}\,n_sn_z)$.
Note this is linear in $n_sn_z$, the size of the dynamic optimal
discretization over $S\times X$. Thompson sampling with
discretization uses one operation that scales as $O(n^2(n_sn_z) + n(n_sn_z)^2 + (n_sn_z)^3)$
in the worst case and to reduce cost special techniques are required e.g. Fourier features,
CG matrix inversion.

%% file: secs/5-Experiments-Neurips.tex
\section{Experiments}
We consider synthetic benchmarks, the three applications described
in Section~\ref{sec:probdef}, and in the SM we also present parallelization results
and global optimization experiments comparing $\KG$ variants
to other methods. We observe that $\KG_{MC}$ performs worse in
the same computation time hence we exclude $\KG_{MC}$ methods
from the conditional experiments.

\begin{figure}[b]
\label{fig:synth}
    \centering
    \includegraphics[width=0.49\textwidth]{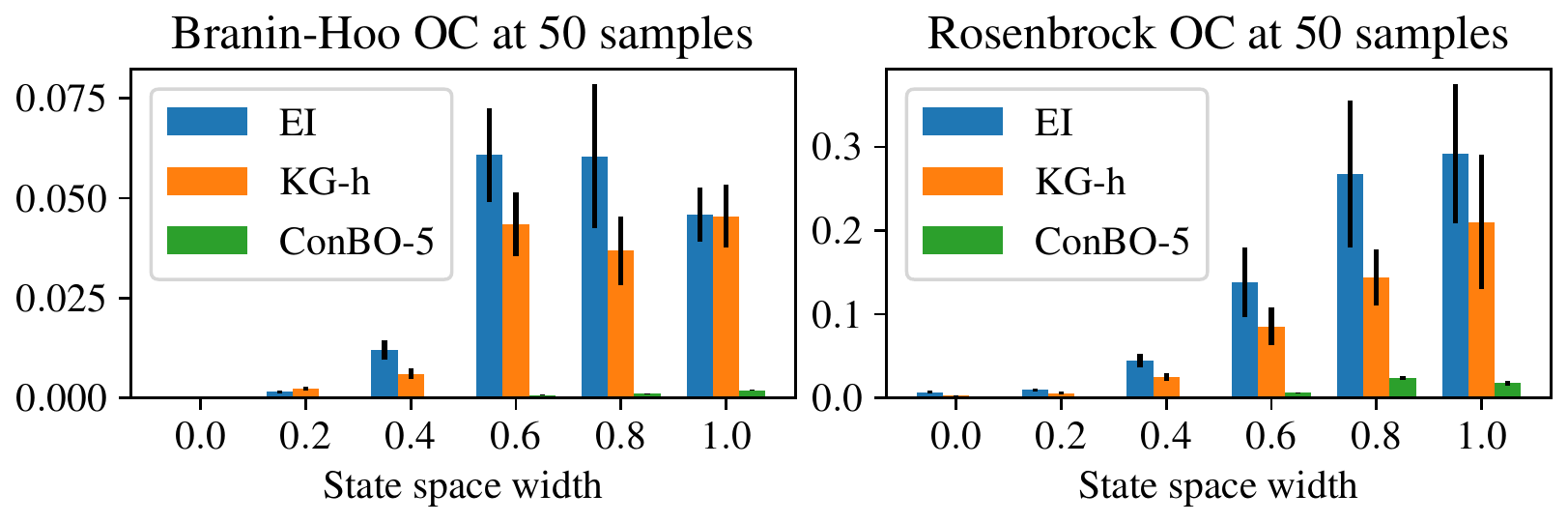}
    \caption{Opportunity cost (simple regret) on two synthetic functions with varying state space width. A width of zero reduces the problem to global optimization for which global methods, EI, $\KG_h$,
    perform well. Increasing state space width to include more states to optimize, 
    the ConBO algorithm performance is consistent while global methods suffer.}
    \label{fig:glob_on_cond}
\end{figure}

\begin{figure*}[t]
    \centering
    \includegraphics[width=\textwidth]{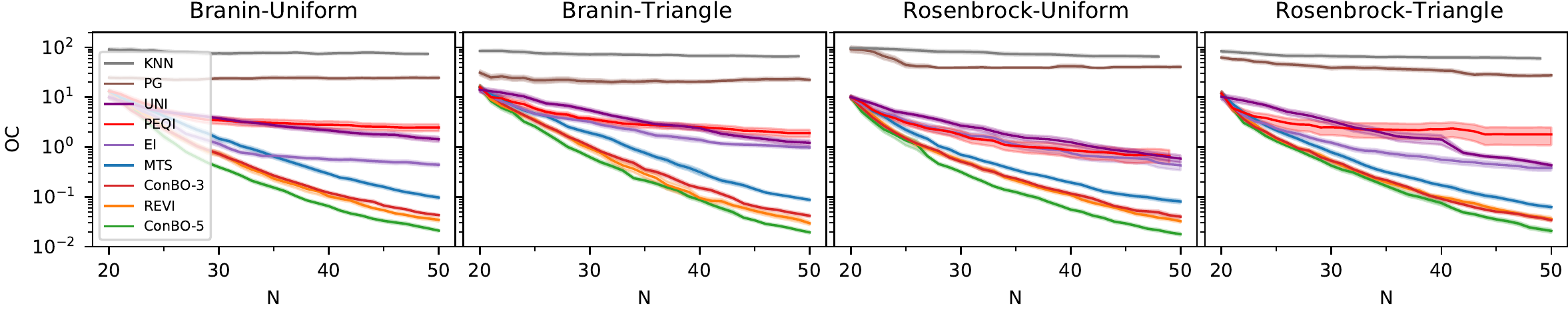}
    \caption{Opportunity Cost across a range of synthetic test problems. 
    The dummy baseline, KNN, is worst in all cases, policy gradient
    is better however the Gaussian process based methods all perform better. 
    UNI and EI are not conditional algorithms yet outperform PEQI. Amongst other
    conditional algorihtms, MTS, REVI and ConBO methods all perform significantly
    better. ConBO-3 is outperformed by ConBO-5 demonstrating
    the improvement with more accurate $\KG_h$.}
    \label{fig:res_synth}
\end{figure*}


\subsection{Synthetic Functions} \label{sec:exp_synth}
We perform low dimensional toy experiments in an ideal setting as a sanity
check where the error due to discretizations will be minimal.
We use the popular Branin-Hoo and Rosenbrock test functions in 2D corresponding
to the (state, action) domain as displayed in Figure~\ref{fig:glob_on_cond}.

\textbf{Global Algorithms on Conditional Problems}
We first modify the synthetic conditional benchmark to construct a range of problems
that vary only by the width of the state space. Zero width corresponds
to a singleton state space (e.g. $S=\{0\}$ for Rosenbrock) so that the problem reduces to global optimization,
whilst a width of one corresponds to the original continuum of states (e.g. $S=[-2, 2]$ for Rosenbrock).
We apply expected improvement and hybrid Knowledge Gradient, these acquisition functions aim to
find the single global maximum over $S \times X$. We also apply ConBO
that tries to find the maximum of each state. All methods use the fitted GP to define the policy,
therefore differences are solely due to the acquisition function.
Results are in Figure~\ref{fig:glob_on_cond}. When state space width
is zero, all methods achieve near zero opportunity
cost (OC), both global and conditional algorithms
find the optimal action of the single state. As state space width
increases to one, there are increasingly more states to optimize.
ConBO consistently achieves near zero total OC over test states. Meanwhile
global methods have increasing OC with state space width, these methods
prioritise sampling high reward states neglecting the optimization
of every state and failing to converge.

\textbf{Conditional Algorithms} Using the same synthetic functions with full state space width, we consider a uniform and a triangular state distribution $\P[s]$. For baselines, we adopt two policy based methods. \textbf{KNN}: randomly collected data, the policy returns the best observed action from 10 nearest neighbor states and serves as a dummy baseline. \textbf{PG}: policy gradient, a parametric quadratic policy $\pi_\theta(s)$ is learnt by maximising observed rewards, data collection samples states from $\P[s]$ then $\epsilon$-greedy actions. For a controlled experiment, all BO methods fit a GP defining the policy and only differ by data collection (using a different policy e.g. recommend global optimum $x$ or single $x$ best on average over states, is clearly inferior and we do not investigate it here). \textbf{UNI}: random data collection, \textbf{PEQI}: given $(s,x)$, computes 0.75 quantile improvement of $(s,x)$ over highest 0.75 quantile within the same state. \textbf{REVI}: discretizes $S\times X$, computes $\KG_d$ for each state. \textbf{MTS}: discretizes $S\times X$, draws a sample output vector, the state with highest improvement over the policy action for the same state is chosen with the corresponding action. \textbf{ConBO-k}: given $(s,x)$, 20 states are importance sampled, $\KG_h$ with $k=3,5$ points is used. \textbf{EI}: expected improvement, optimization over $S\times X$ treating all inputs as actions, the same as in the previous section.

Results are in Figure~\ref{fig:res_synth}. 
Policy based methods KNN and PG consistently perform worse than
the  Gaussian process methods (which may be viewed as value based
as they predict the reward of any state-action pair). Surprisingly, the conditional
BO algorithm PEQI performs similarly to UNI and much worse than EI.
In all cases, the conditional methods outperform all non-conditional methods.



\begin{figure*}[t]
    \centering
    \includegraphics[width=\textwidth]{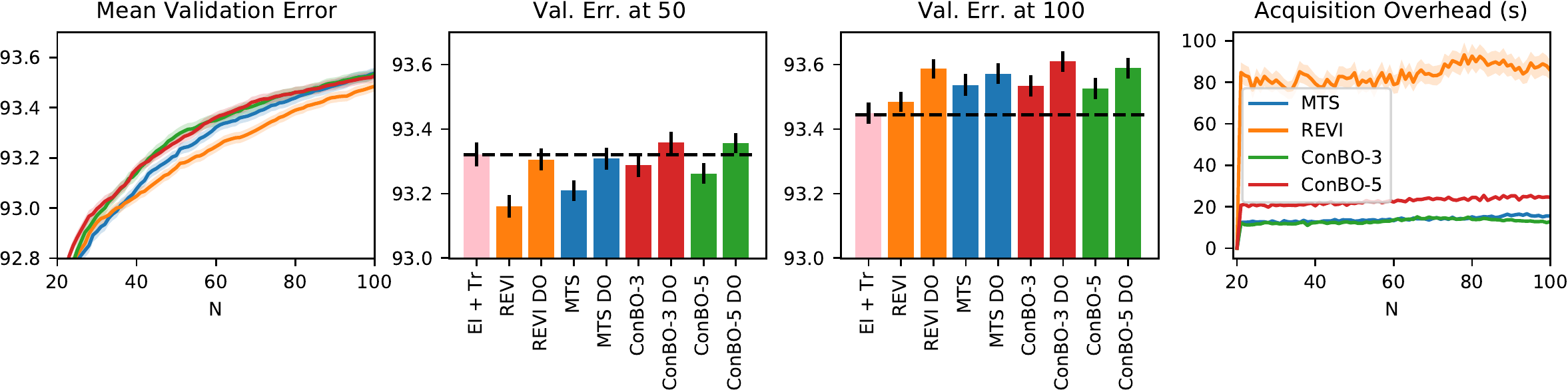}
    \caption{Left: validation error. Centre-left: validation error after 50 samples. 
    Centre-right: validation error after 100 samples. Right: algorithm overhead in seconds.
    After 50 samples, none of the multi-task models outperform the baseline, EI + Tr (dashed line) 
    suggesting all datasets can use similar hyperparameters.
    For the larger budget 100, all models outperform the baseline by $0.1\%$
    suggesting that for more fine-tuning, each dataset requires different
    hyperaparameters. In all cases, performing data optimization significantly
    increases performance.}
    \label{fig:res_CNN}
\end{figure*}

\subsection{CNN Training Hyperparameters}
We apply MTS, REVI, and ConBO variants and we adopt the recently
proposed kernel used for BO with Common Random Numbers~\citep{pearce2019bayesian},
\begin{align*}
k((s,x), (s',x')) &= \\
\sigma_0^2M(x,x'; &\underline{l}) + \delta_{s's}( \sigma_1^2M(x,x';\underline{l}) + \sigma^2_3).
\end{align*}
where $M(x,x';\underline{l})$ is a Mat\'ern $\frac{5}{2}$ kernel with length scales $\underline{l}$. The first term models
a common trend function across all states and the second term models how each state independently
differs from the trend. The differences are composed of another Mat\'ern and the constant kernel
to model a global offset e.g. one dataset may have universally lower validation error.
This kernel has far fewer parameters than a full multi-task product kernel, it is easy 
to fit and scales to arbitrary number of states (or datasets) without adding extra parameters.

In this problem setting, learning hyperparameters over similar
datasets, one may expect that the optimal hyperparameters would
be the same for all datasets. Therefore, as a baseline we apply
EI to learn the hyperparameters of the first dataset (state 1). We then evaluate the objective function (validation error) on the rest of the datasets using the best observed hyperparameters from
dataset 1, we refer to this as \textbf{EI + Tr}ansfer.

\textbf{Policy Optimization versus Data Optimization:} in discrete state settings where the number of states 
is much lower than the sampling budget $|S| \ll N$, a user may desire to find good observed (stochastic)
values $\max y$ for each state instead of finding $\pi(s)=\argmax_x \E[f(s,x)]$. Here, we refer to the
former as \emph{data optimization} (DO) and the latter as \emph{policy optimization} (PO). For stochastic
simulation, PO methods are used to find actions that have
reproducible average rewards as the actions will be deployed for real world
use. In neural network training we mainly intend to find a network with good validation error
regardless of the hyperparameters as the network will be deployed
for real world use, hence DO is preferred. Past work~\citep{pearce2018continuous}
showed that a PO method can be used to warm start a DO method by performing
PO and reserving samples at the end to allocate one per
state with action determined by EI (which is a DO algorithm). This was shown to
significantly outperform using a DO algorithm for all
samples. We apply this ``DO trick'' to all algorithms in this experiment.


Results are shown in Figure~\ref{fig:res_CNN}. For the medium budget of 50 samples, ConBO
performs best of the standard algorithms yet it is still is worse than the baseline.
Applying the final round of DO improves all results to match the baseline.
For the large budget of 100 samples, all methods outperform the baseline suggesting dataset
specific fine-tuning of hyperparameters is required to achieve best results. Again, DO provides
a significant boost to performance of all methods.

Gaussian process kernel parameter learning required approximately 2--5 seconds
using Tensorflow. In figure \ref{fig:res_CNN} (right) we show the run time
of each algorithm excluding model fitting and network training, purely
acquisition function optimization time. MTS and ConBO-3 are quickest
while Conbo-5 increases linearly over ConBO-3 and REVI takes much longer.

In future work we intend to investigate more hyperparameter specific extensions such as
early stopping, variable training set size, pretraining etc. In these preliminary
experiments we consider the base case and treat the training process as a black box.

    

\subsection{Ambulances and Warehouses}

\begin{figure*}[t]
    \centering
    \includegraphics[width=\textwidth]{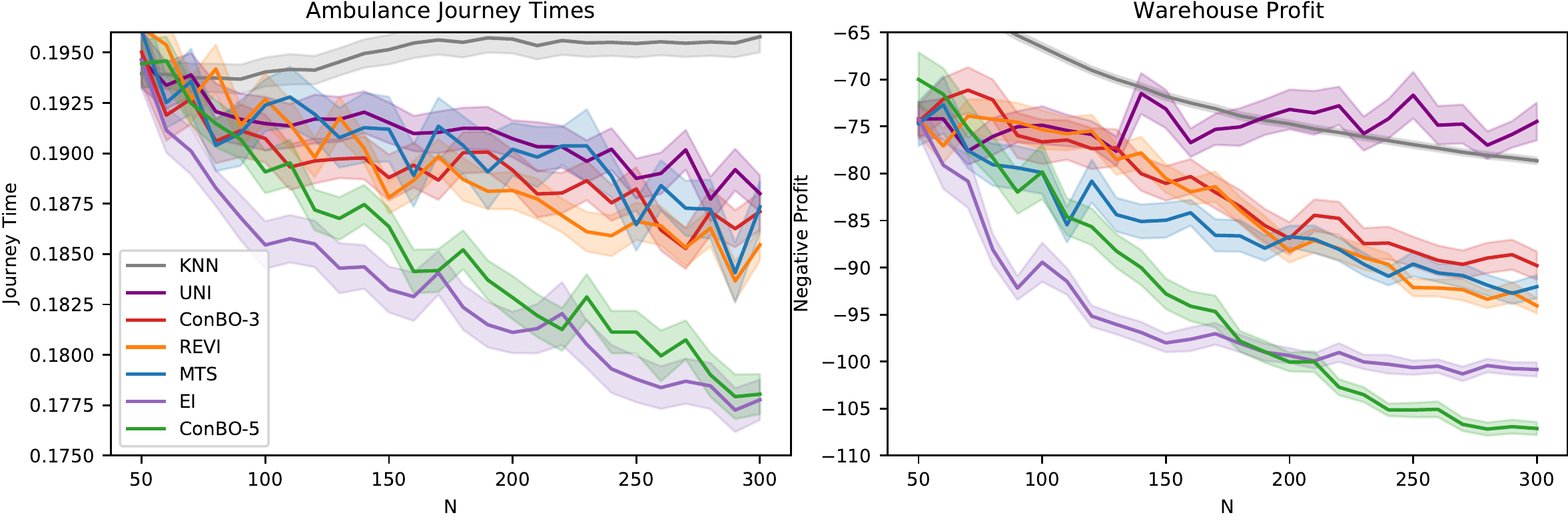}
    \caption{Left: average journey times across a range of cities. Right : average profit 
    across a range of warehouses. ConBO-5 and EI perform best on these benchmarks.}
    \label{fig:res_amb_atox}
\end{figure*}

We apply all the methods from Section \ref{sec:exp_synth} to two
benchmarks from the {\tt www.SimOpt.org} library for simulation
optimization problems. The ambulance problem (AMB) consists of
a range of cities with different population distributions and one
must optimize ambulance base locations for each city. The Assemble-to-order
problem (ATO) consists of a range of warehouses that face different demand
and for each warehouse the target stock level must be optimized.

Results are shown in Figure~\ref{fig:res_amb_atox}. Of the policy based methods,
PG performs poorly and does not show on the plots whilst KNN performs poorly on AMB
and performs well on ATO suggesting AMB is a more difficult problem. Of the GP based
methods, EI performs well. Although it is not a conditional algorithm we include it to 
highlight that sometimes the simplest idea can also work, note that learning
the global optimum $\argmax_{s,x}f(s,x)$ also learns the corresponding part of
the policy for the ``optimum'' state $\pi(s^*)$. However we emphasize that this behaviour will not
generalize to other problems as shown by the synthetic results and the premature
convergence on the ATO problem. Of the conditional methods, MTS, REVI, and ConBO-3 all perform
similarly, either slightly (AMB) or largely (ATO) outperforming UNI. ConBO-5 uses a
more accurate acquisition function and is the only method that
learns a good policy on both problems. This implies that these problems are more difficult
than the synthetics and CNN and truly stress test conditional algorithms.

\textbf{Parallelization}
Unique to the conditional setting, one may execute function calls on different states
in parallel as long as the states are far apart and there is little interaction. In SM we use sequential batch 
construction by penalization~\citep{gonzalez2016batch} for ConBO and observe almost perfect linear scaling
in AMB and mixed results on ATO with almost no added computation overhead.


%% file: secs/7-Conclucsion-Neurips.tex
\section{Conclusion}
We propose ConBO, a simple theoretically grounded algorithm framework for conditional Bayesian optimization.
We investigate this method on a range of problems and in all benchmarks 
compared to state of the art methods
ConBO-5 was either best or joint best algorithm providing the most reliable
consistent performance, with ConBO-3 a cheaper alternative. We also emphasise
the caution required with additive kernels and the conflict of 
data optimization versus policy optimization.

%% file: sm_secs/0.5theory.tex
\setcounter{theorem}{0}
\section{Theoretical Results}
We restate the theorems from the main paper and provide each proof. Firstly, in Theorem~\ref{th:one} we 
show that that ConBO with knowledge gradient myopically maximises \emph{Value of Information}
in a Bayesian decision theoretic framework. In Theorem~\ref{th:two} we show that in discrete settings 
ConBO will sample all pairs infinite often. Finally, in Theorem~\ref{th:three} we prove conditions for
hybrid KG to satisfy the results of Theorems 1 and 2.

\begin{theorem}\label{th:one}
Let $(s^*, x^*) = \argmax \CONBO(s, x)$ be a point chosen for sampling. $(s^*, x^*)$ is also the point that
maximises the myopic Value of Information, the increase in predicted policy reward.
\end{theorem}

\textit{Proof of Theorem 1}
Given all the information available at time $n$, $\tilde X^n, Y^n$ and the model $\mu^n(s,s)$, $k^n(s,x,s'x')$,
for any given state and action $s, x$, and a given realization of the true reward function $f()$,
in a Bayesian decision theoretic framework, the loss is given by the output of the function 
$$\text{Loss}(s,x)=-f(s,x),$$
the expected loss is the risk function
$$\text{Risk}(s,x)=\E[\text{Loss}(s,x)|\tilde X^n, Y^n]=\E[-f(s,x)|\tilde X^n, Y^n]=-\mu^n(s,x).$$
For convenience we assume conditioning on $\tilde X^n, Y^n$ for the remaining equations.
The optimal action minimizes risk 
$$x^{optimal} = \argmin_x \text{Risk}(s,x) = \argmin_x -\mu^n(s,x).$$
Alternatively, $\pi(s)=\argmax_x \mu^n(s,x)$ is the Bayesian decision
theoretic optimal action given all data available at time $n$. The total risk
is the risk of optimal actions is for all states, or the risk of the policy
$$
\text{Total Risk}(n) = - \int_s \max_x \mu^n(s,x) \P[s]ds
$$
which is the negative of the models best prediction of true reward given data up to time $n$
where we have made $n$ an explicit argument for convenience.
Next assume we are able to collect more data to update the model, choose $(s,x)\n1$ and observe $y\n1$.
The myopic \textit{Value of Information} is defined as the data that minimizes future risk
\begin{eqnarray}
\VoI((s,x)\n1) &=& -\E_{y\n1}[\text{Total Risk}(n+1)|(s,x)\n1]  \\
\end{eqnarray}
Note that that $\argmax \VoI((s,x)\n1)$ is not affected by adding terms that do not depend on $(s,x)\n1$.
Thus we may subtract the current $\text{Total Risk}(n)$. Finally, the difference between risks simplifies to
\begin{eqnarray}
&&\E_{y\n1}\left[ \int_s \max_x \mu\n1(s,x) \P[s]ds\bigg|(s,x)\n1\right] - \int_{s'} \max_x \mu^n(s',x) \P[s]ds'\\
&=& \E_{y\n1}\left[ \int_s \max_x \mu\n1(s,x) - \max_x \mu^n(s,x) \P[s]ds  \bigg|(s,x)\n1\right] \\
&=& \int_s \E_{y\n1}\left[  \max_x \mu\n1(s,x)|(s,x)\n1\right] - \max_x \mu^n(s,x)  \P[s]ds \\
&=& \int_s \KG_c(s; (s,x)\n1) \P[s]ds \\
&=& \text{ConBO}((s,x)\n1)
\end{eqnarray}
Therefore $\argmax \VoI((s,x)\n1) = \argmax \text{ConBO}((s,x)\n1)$.\qed

\begin{theorem}\label{th:two}
Let $S$ and $X$ be finite sets and $N$ the budget to be sequentially
allocated by $\CONBO$. Let $n(s,x, N)$ be the number of samples
allocated to point $s,x$ within budget $N$. Then for all $(s,x) \in S \times X$
we have that $\lim_{N\to \infty} n(s,x, N)=\infty.$
\end{theorem}

We require some intermediate results, firstly ConBO is non-negative.
\begin{lemma} Let $(s,x) \in S\times X$, then 
$\CONBO((s,x)\n1) \geq 0.$
\end{lemma}

\textit{Proof of Lemma 1}
\begin{eqnarray}
\CONBO((s,x)\n1) 
&=& \sum_{s'} \E_Z[\max_{x'} \mu^n(s',x') + \sigt(s',x'; (s,x)\n1)Z] - \max_{x''} \mu^n(s',x'') \\
&\geq& \sum_{s'}\E_Z[  \mu^n(s',\pi^n(s')) + \sigt(s',\pi^n(s'); (s,x)\n1)Z] - \max_{x''} \mu^n(s',x'') \\
&=& \sum_{s'} \max_{x'}\mu^n(s',x') + \sigt(s',\pi^n(s'); (s,x)\n1)\E_Z[Z] - \max_{x''} \mu^n(s',x'') \\
&=& \sum_{s'} \max_{x'}\mu^n(s',x') - \max_{x''} \mu^n(s',x'') \\
&=& 0
\end{eqnarray}\qed

Secondly, we require that ConBO$(s,x)$ reduces to zero for an infinitely sampled pair. Note that for deterministic $f(s,x)$,
the result simplifies to ConBO$(s,x)$ is zero for any sampled pair.
\begin{lemma} Let $(s,x)\n1 \in S\times X$ with $n(s,x)=\infty$, then 
$\CONBO((s,x)\n1) = 0.$
\end{lemma}

\textit{Proof of Lemma 2}
Given infinitely many finite variance observations of $f(s,x)$, we have that $\mu^n(s,x)=\E[f(s,x)]$
and posterior variance is zero $k^n(s,x,s,x)=0$. By the positive definiteness of the kernel we also have that
$k^n(s,x, s',x')=0$ for all $(s',x')\n1 \in S\times X$ (see \cite{PearceBranke1} Lemma 3). It follows that
$\sigt(s',x';(s,x)\n1)=0$ for all $(s',x') \in S\times X$ and thus
\begin{eqnarray}
\KG_c(s'; (s,x)\n1) 
&=& \E_Z[ \max_{x'}\mu^n(s', x') + \sigt(s', x'; (s,x)\n1)Z] - \max_{x''}\mu^n(s', x'') \\
&=& \E_Z[ \max_{x'}\mu^n(s, x') + 0\cdot Z] - \max_{x''}\mu^n(s', x'') \\
&=&  \max_{x'}\mu^n(s', x') - \max_{x''}\mu^n(s', x'') \\
&=& 0
\end{eqnarray}
and therefore ConBO$((s,x)\n1) = \int_s \, 0 \, \P[s]ds=0$.\qed


Thirdly, we require the inverse of Lemma 2, that points for which ConBO$(s,x)>0$ must have 
non-zero variance $k^n(s,x,s,x)>0$ (and therefore cannot be infinitely sampled).

\begin{lemma}
Let $(s,x)\n1 \in S\times X$ be a point for which ConBO$((s,x)\n1)>0$, then
$n(s,x)< \infty$.
\end{lemma}
\textit{Proof of Lemma 3}
$\ConBO((x,s)\n1)>0$ implies that there exists an $s\in S$' such that
$KG_c(s';(s,x)\n1)>0$. By the contrapositive of Lemma 3 in \cite{poloczek2017multi}, 
we must have that $k^n(s',x',(s,x)\n1)$
is \emph{not} a constant function of $x'$. If $(s,x)\n1$ is infinitely sampled,
then $k^n(s',x',(s,x)\n1)$ is a constant function of $x'$, thus $(s,x)\n1$ is not infinitely sampled.\qed

Finally, combining the previous Lemmas we can complete the proof. 

\textit{Proof of Theorem 2}
By Lemmas 1 and 2, any infinitely sampled points become minima of the function $\ConBO(s,x)$.
By construction, the ConBO algorithm choose points at maxima $(s,x)\n1=\argmax \ConBO(s,x)$. Thus in the infinite budget limit, 
we have $\ConBO(s,x)=0$ for all $(s,x)\in S\times X$ by the contrapositive of Lemma 3 we have that $n(s,x)=\infty$ for all points.\qed

\begin{theorem}\label{th:three}
Let $n_z\geq 2$ and let $\underline{Z} = \{Z_j|j=1,...,n_z\}$. If $0 \in \underline{Z}$
then $KG_h(x)\geq 0$ for all $x\in X$ and if $x$ is sampled infinitely often $\KG_h(x) = 0$.
\end{theorem}
\textit{Proof of Theorem 3}
First consider the base case $n_z=2$. Let $\underline{Z}=\{0, Z_2\}$ and given
$x\n1$, let $X^*=\{x^*_1, x^*_2\}$ be the optimal discretization as found by Algorithm~\ref{alg:hybridKG}.
Then $x^*_1 = \argmax \mu^n(x) + \sigt(x;x\n1)\cdot 0 = \argmax \mu^n(x)$ and therefore
$\mu^n(x^*_1) = \max_x\mu^n(x)$. Let $\mu^* = \mu^n(X^*)$ and $\sigt^* = \sigt(X^*, x\n1)$.
Then we have that
\begin{eqnarray}
KG_h(x\n1) 
&=& \E_Z\left[\max\{ \mu^*_1 + \sigt^*_1Z, \mu^*_2 + \sigt^*_2Z \} \right] - \max_x\mu^n(x) \\
&=& \E_Z\left[\max\{ \max_x\mu^n(x) + \sigt^*_1Z, \mu^*_2 + \sigt^*_2Z \} \right] - \max_x\mu^n(x) \\
&=& \E_Z\left[\max\{ \sigt^*_1Z, \mu^*_2- \max_x\mu^n(x) + \sigt^*_2Z \} \right]  \\
&\geq& \max\left\{\E_Z\left[ \sigt^*_1Z\right], \E_Z\left[\mu^*_2- \max_x\mu^n(x) + \sigt^*_2Z \right] \right\}  \\
&=& \max\left\{0, \mu^*_2- \max_x\mu^n(x) \right\}  \\
&=& 0
\end{eqnarray}
where we Jensen's inequality in the penultimate line and we use that $\mu^*_2< \max_x\mu^n(x)$ in the final line.
The result extends to the case for $n_z>2$ trivially.
The proof for $\KG_h(x)=0$ at infinitely sampled points follows the proof of Lemma 2.\qed

%% file: sm_secs/0-hybrid-KG.tex
\section{Computing ConBO and Hybrid Knowledge Gradient}

\begin{algorithm}
\caption{Computing ConBO. The algorithm requires a new point, discretization sizes, past data
and posterior GP functions and an optimizer.}
\begin{algorithmic}
\Require $\tilde x\n1$, $n_s$, $n_z$, $\tilde X^n$, $Y^n$, $\mu^n(s,x)$, $k^n(s,x,s'x')$, $\sigma_\epsilon^2$, {\tt Optimizer()}
\State Precompute and cache $\left( k^0(\tilde X^n, \tilde X^n) + \sigma_\epsilon^2I \right)^{-1}k^0(\tilde X^n, \tilde x\n1)$
\State $C \gets 0$
\For{$i$ \textbf{in} $1,..,n_s$}
    \State $s_i\sim N(s_i | s\n1, \text{diag}(l_s^2))$
    \State $KG_i \gets \text{Algorithm}$\ref{alg:hybridKG}$(\tilde x\n1, s_i,....)$
    \State $w_i \gets \P[s_i] / N(s_i | s\n1, \text{diag}(l_s^2))$
    \State $C \gets C + w_i  KG_i / n_s$
\EndFor \\
\Return $C$
\end{algorithmic}
\end{algorithm}

\begin{algorithm}
\caption{Computing Hybrid Knowledge Gradient. The algorithm requires a new point, a state, a discretization
size, past data, posterior GP functions and an optimizer.\label{alg:hybridKG}}
\begin{algorithmic}
\Require  $\tilde x\n1$, $s$, $n_z$, $\tilde X^n$, $Y^n$, $\mu^n(s,x)$, $k^n(s,x,s'x')$, $\sigma_\epsilon^2$, {\tt Optimizer()}
\State $\tilde X\n1 \gets \tilde X^n \cup \{\tilde x\n1\}$
\State $X^* \gets \{\}$
\State Precompute and cache $\left( k^0(\tilde X^n, \tilde X^n) + \sigma_\epsilon^2I \right)^{-1}k^0(\tilde X^n, \tilde x\n1)$
\For{$j$ \textbf{in} $1,..,n_z$}
    \State $Z_j \gets \Phi^{-1}\left(\frac{2j-1}{2n_z}\right)$
    \State $\tilde{Y}\n1_j $ from Equation~\ref{eqn:Yn1}
    \State $\mu_j\n1(s, x) \gets \mu^0(s, x) + k^0(s, x,\tilde X\n1)\tilde Y\n1_j$
    \State $x^*_j \gets \argmax_x \mu_j\n1(s, x)$ using {\tt Optimizer()}
    \State $X^* \gets X^* \cup \{x^*_j\}$
\EndFor \\
$\underline{\mu} \gets \mu^n(s, X^*)$\\
$\underline{\sigma} \gets \frac{k^n((s,X^*), \tilde x\n1)}{\sqrt{k^n(x\n1, x\n1)+\sigma_\epsilon^2}}$ \\
    KG $\gets$ Algorithm\ref{alg:KGdisc}$(\underline{\mu}, \underline{\sigma})$ \\
\Return KG
\end{algorithmic}
\end{algorithm}

\begin{algorithm}[!h] \caption{Knowledge Gradient by discretization. This algorithm takes as input a set of linear
functions parameterised by a vector of intercepts $\mul$ and a vector of
gradients $\sigl$. It then computes the intersections
of the piece-wise linear epigraph (ceiling) of the functions and 
the expectation of the output of the function given Gaussian input. 
Vector indices are assumed to start from 0.
\label{alg:KGdisc}}
\begin{algorithmic}
\Require $\mul$, $\sigl \in \R^{n_A}$
\State $O \gets \text{order}(\sigl)$ \,\,\,\quad \quad \quad                \,\,\# get sorting indices of increasing $\sigl$
\State $\mul \gets \mul[O]$, $\sigl \gets \sigl[O]$ \quad                   \# arrange elements
\State $I\gets[0,1]$ \quad\quad\quad\quad\quad                              \,\,\,    \# indices of elements in the epigraph
\State $\Zt \gets [-\infty, \frac{\mu_0 - \mu_1}{\sigma_1 - \sigma_0}    ]$ \,\quad\quad\# z-scores of intersections on the epigraph
\For{$i=2$ \textbf{to} $n_z-1$}
	\State ($\star$) 
	\State $j\gets last(I)$ 
	\State $z\gets \frac{\mu_i - \mu_j}{\sigma_j - \sigma_i} $
	\If {$z<last(\Zt)$}
		\State Delete last element of $I$ and of $\Zt$
		\State Return to ($\star$)
	\EndIf
	\State Add $i$ to end of $I$ and $z$ to $\Zt$
\EndFor
\State $\Zt\gets [\Zt,\infty]$
\State $\underline{A} \gets \phi(\Zt[1:]) - \phi(\Zt[:-1])$ \quad\quad\quad \# assuming python indexing
\State $\underline{B} \gets \Phi(\Zt[1:]) - \Phi(\Zt[:-1])$ 
\State $\KG \gets \underline{B}^T\mul[I] - \underline{A}^T\sigl[I] - \max \mul$ \,\, \# compute expectation
\State \Return KG
\end{algorithmic}
\end{algorithm}

\subsection{Deriving One-Step Look-ahead Posterior Mean $\mu^{n+1}(s,x)$}
At iteration $n$ during optimization, let the training inputs be 
$\tXn=\left( (s^1, x^1),...,(s^n, x^n)\right)$ and
the training outputs $Y^n = (y^1,...,y^n)$. Given a prior mean and kernels functions, $\mu^0(s,x):S\times X\to \R$ and $k^0(s,x,s',x'): S\times X \times S\ \times X \to \R$.
Finally let the new sample point be $(s,x)\n1=\txn1$.

Updating the mean function with data from the $0^{th}$ step to $n^{th}$ step is given by
\begin{eqnarray}
    \mu^n(s,x) &=& \mu^0(s,x) + k^0(s,x, \tXn) \underbrace{K^{-1}\left(Y^n - \mu^0(\tXn)\right )}_{\text{define }\tilde Y^n}\\
    &=& \mu^0(s,x) + k^0(s,x, \tXn)\tilde Y^n \label{eq:factorized_post_mean}
\end{eqnarray}
where $K=k^0(\tXn, \tXn)+\sigma_\epsilon^2I$. $\mu^n(s,x)$ may also be written as a weighted average of a modified $\tilde Y^n\in\R^n$ vector as defined above. Computing the new posterior mean reduces to augmenting $\tXn \to \tilde X\n1$ with $\txn1$ and
$Y^n\to Y\n1$ then computing the new $\tilde Y\n1\in \R\n1$. Let $Z$ be the z-score of $y\n1$ on its predictive distribution, then 
\begin{equation} \label{eqn:Yn1}
    \tilde Y\n1 = 
    \begin{bmatrix}
    \tilde Y^n \\
    0
    \end{bmatrix}
    +\frac{Z}{\sqrt{ k^n(\txn1, \txn1) + \sigma_\epsilon^2}}
    \begin{bmatrix}
    -K^{-1}k^0(\tXn, \txn1) \\
    1
    \end{bmatrix}
\end{equation}
and the above expression may be used directly in Algorithm 1 with sampled $Z_j\sim N(0, 1)$.
This is derived by a simple change of indices from $0\to n$ and $n \to n+1$, yields the one-step updated posterior mean
\begin{equation}
    \mu\n1(s,x) = \mu^n(s,x) + \frac{k^n(s,x, \txn1)}{k^n(\txn1, \txn1)+\sigma_\epsilon^2}
    \left(y\n1 - \mu^n(\txn1)\right ).
\end{equation}
which contains the random $y\n1$. This may be factorized as follows:
\begin{eqnarray}
    \mu\n1(s,x) &=& \mu^n(s,x) + 
    k^n(s,x, \txn1)\frac{1}{\underbrace{\sqrt{ k^n(\txn1, \txn1)+\sigma_\epsilon^2}
    }_{\text{standard deviation of $y\n1$}}}
    \underbrace{\frac{\left(y\n1 - \mu^n(\txn1)\right )}
    {\sqrt{ k^n(\txn1, \txn1)+\sigma_\epsilon^2}}
    }_{\text{Z-score of $y\n1$}} \\
    &=& \mu^n(s,x) + k^n(s,x, \txn1)\frac{1}{\sigma^n_{y\n1}(\txn1)}Z \label{eq:factorized_new_post_mean} \\
    &=& \mu^n(s,x) + \tilde\sigma(s,x, \txn1) Z
\end{eqnarray}
where the left factor is a deterministic and the right factor is
the (at time $n$) stochastic Z-score of the new $y\n1$ value. This is clear by
noting that the predictive distribution of the new output $y\n1$
\begin{equation}
    \P[y\n1|\txn1, \tXn, Y^n] = N(\mu^n(\txn1), k^n(\txn1, \txn1)+\sigma_\epsilon^2).
\end{equation}
as a result, to sample new posterior mean functions, we may simply sample
$Z\sim N(0,1)$ values and compute Equation \ref{eq:factorized_new_post_mean}.
However, this results in a quadratic cost per call to sampled poster mean function
as both $k^n(s,x,\txn1)$ and $\sigma^n_{y\n1}(\txn1)$ have $O(n^3)$quadratic cost. This
can be easily reduced to linear instead as we now show.

We next focus on the first factor $k^n(s,c,\txn1)$ which may also be factorized
\begin{eqnarray}
    k^n(s,x,\txn1) 
    &=& k^0(s,x,\txn1) - k^0(s,x,\tXn) K^{-1}k^0(\tXn, \txn1) \\
    &=& \label{eq:factorized_post_cov}
    \underbrace{\begin{bmatrix}
        k^0(s,x, \tXn), k^0(s,x,\txn1)
    \end{bmatrix}}_{k^0(s,x, \tilde X\n1)}
    \begin{bmatrix}
    -K^{-1}k^0(\tXn, \txn1) \\
    1
    \end{bmatrix}.
\end{eqnarray}
Combining Equations \ref{eq:factorized_post_mean} and \ref{eq:factorized_post_cov}
yields the following formula
\begin{eqnarray}
    \mu\n1(s,x) &=& \mu^0(s,x) + k^0(s,x,\tilde X\n1)\left( 
    \begin{bmatrix}
    \tilde Y^n \\
    0
    \end{bmatrix}
    +\frac{Z}{\sigma^n_{y\n1}(\txn1)}
    \begin{bmatrix}
    -K^{-1}k^0(\tXn, \txn1) \\
    1
    \end{bmatrix}
    \right) \\
    &=& \mu^0(s,x) + k^0(s,x,\tilde X\n1)\tilde Y\n1. \label{eq:cheap_new_post_mean}
\end{eqnarray}
The quantity $\tilde Y^n$ is pre-computed at the start of the algorithm iteration,
the quantity $-K^{-1}k^0(\tXn, \txn1)$ has quadratic cost and can be
computed once and used again for $\sigma^n_{y\n1}(\txn1)$. Then, sampling
posterior mean functions reduces to sampling $n_y$ values $z_1,...,z_{n_y}\sim N(0,1)$
and for each value computing $\tilde Y\n1_1,....,\tilde Y\n1_{n_y}$. Then
each sampled posterior mean is just the weighted average given by Equation \ref{eq:cheap_new_post_mean}.


\subsection{Choice of {\tt Optimizer()}}
Since evaluating Equation \ref{eq:cheap_new_post_mean} for many points
is simply a matrix multiplication, random search is cheap to evaluate in parallel.
After random search, the gradient of Equation \ref{eq:cheap_new_post_mean} with respect to 
$(s,x)$ is easily computed, and starting from the best random search point,
gradient ascent over $x\in X$ can be used to
find the optimal action. This varies with kernel choice and application, we describe our
settings in Section~\ref{sec:conbo_settings}.

We start with a set of sampled means $\mu_1\n1(s,x),....,\mu_{n_z}\n1(s,x)$ and a set of
sampled states $s_1,...,s_{n_s}$. For each state $s_i$, the set of $n_z$ 
optimal actions $X_{d, s_i}$ is found by optimizing the $n_z$ posterior means
$$
X_{d, s_i} = \bigcup_{j=1}^{n_z} \left\{\underset{x}{\text{argmax}} \,\mu_j\n1(s_i, x)\right\}
$$

Finally, for each point in the $\{s_i\}\times X_{d,s_i} = \tilde X_{d, s_i}$ (state, action) discretization,
we evaluate two quantities, firstly the vector of current posterior means $\underline\mu_{s_i}\in \R^{n_z}$,
\begin{eqnarray}
\underline\mu_{s_i} &=& \mu^n(\tilde X_{d, s_i}) \\
&=& \mu^0(\tilde X_{d, s_i}) + k^0(\tilde X_{d, s_i}, \tXn)\tilde Y^n
\end{eqnarray}
and the vector of additive updates $\underline\sigma_{s_i} \in \R^{n_z}$,
\begin{eqnarray}
\underline\sigma_{s_i}  &=& \frac{k^n(\tilde X_{d, s_i}, \txn1)}{\sigma^n_{y\n1}(\txn1)} \\
&=& k^0(\tilde X_{d, s_i}, \tilde X\n1)
    \begin{bmatrix}
    -K^{-1}k^0(\tXn, \txn1) \\
    1
    \end{bmatrix}\frac{1}{\sigma^n_{y\n1}(\txn1)}.
\end{eqnarray}
These two vectors $\underline\mu_{s_i}$ and $\underline\sigma_{s_i}$  are both differentiable $\nabla_{\txn1}\underline\mu_{s_i}$ and $\nabla_{\txn1}\underline\sigma_{s_i}$ 
and they are used to analytically compute the peicewise-linear 
$\E_Z\left[\max \left( \underline\mu_{s_i} + Z\underline\sigma_{s_i}\right)\right]- \max \underline\mu_{s_i} $
which is also differentiable. Thus assuming fixed $\tilde X_{d, s_1},...,\tilde X_{d, s_{n_s}}$, approximate gradients
are computed and can be used in any stochastic gradient ascent optimizer.

%% file: sm_secs/2-Experiment_settings.tex
\section{Implementation Details}
\subsection{REVI}
At iteration $n$ of the algorithm, we used a discretization of size $n_{disc} = 2n$, 
split equally amongst actions and states $n_s = n_x = \ceil{\sqrt{n_{disc}}}$.
States are sampled from $\P[s]$ and actions are sampled as a latin hypercube over $X$. The acquisition function is optimized by 100 points of random search over $S\times X$ followed by Nelder-Mead ascent starting form the best 20 points in the random search phase.

\subsection{MTS}
We use a target discretization size of $n_{disc}=3000$.
Given $d_s$ states dimensions and $d_x$ action dimensions, we sampled states uniformly, the number of sampled states is given by $n_s = \ceil{(n_{disc})^{d_s/(d_s+d_x)}}$
and the number of actions per state is $n_x=\ceil{n_{disc} / n_s}$ such that $n_s*n_x\approx n_{disc}$. This way the discretization over all states and action dimensions is roughly constant.
For each sampled state $s_i$, $n_x$ actions are generated in three ways. Firstly, the policy is evaluated $x^{\pi}_{s_i}=\pi^n(s_i)$, we generate 40 actions around this policy action. Secondly, we take the 10 nearest neighbor states from the training set, and the points with the 4 largest $y$ values are added to the discretization set with randomly generated neighbors. Finally, remaining actions in the $n_x$ budget come from uniform random sampling over $X$. Each $s_i$ has a bespoke action discretization. Sampled functions are drawn using the python numpy random normal generate function.

\subsection{ConBO}\label{sec:conbo_settings}
Each sampled posterior mean function was optimized in two steps. For a given state $s_i$, firstly, the action discretization used by MTS, reduced to 40 points in total was used in parallel random search. The best point was then used in conjugate-gradient ascent for 20 steps.

For optimizing sampled posterior mean functions for which $z_i=0$, that is $\mu\n1(s,x)=\mu^n(s,x)$, this given by the policy $x^{\pi}_{s_i} = \pi^n(s_i)$. Since the same, or very similar states, may be used multiple times for different ConBO$((s,x)\n1)$
calls, we may use caching to avoid such repeated policy computation. Whenever the policy is queried for the optimal action for a given state, the final (state, action) pair are stored in a lookup table. Any future calls to the policy function with state $s_j$ can check the lookup table and if very similar states exists use the same action, if a somewhat similar state exists, re-optimize the action, if no similar states exist perform a full optimization as above.

In our experiments, the cache of stored policy calls is wiped clean before any testing, ConBO is not given an unfair advantage at test time. In practical applications, this need not be the case.

\subsection{Policy Gradient}
The (stochastic) policy was a Gaussian with mean that is a
quadratic function of the state and constant variance,
$$
\pi_\theta(s) = \P_\theta[x|s] = N(x | s^TAs + Bs + C, 0.2^2 I )
$$
where $A\in \R^{d_X\times d_S \times d_S}$, $B\in \R^{d_X\times d_S}$ 
and $C \in \R^{d_X} $ and $\theta = \{A, B, C\}$.
Given a dataset $\{(s,x,y)^i\}_{i=1}^n$, we first rescaled $s$ and $x$
values to the hypercube, and y-values were standardized to have mean 0 and variance 1.
First we used kernel regression to learn a baseline value
$$
V(s) =\frac{ \sum_i k(s, s^i)y^i }{\sum_i k(s, s^i)}
$$
where $k(s, s') = \exp( -0.5 (s-s')^2 / 0.2^2)$. 
The parameters $\theta$ are found by optimizing the expected advantage
$$
\text{Expected advantage} = \sum_i \P_\theta[x^i | s^i] (y^i - V(s^i)).
$$
At test time, given a state, the mean action
is computed from the policy (accounting for rescaling to hypercube and back) and 
recommended for use.

\subsection{CIFAR-10 Hyperparameter Experiment}
\textbf{Parameter Space:}
\begin{itemize}
    \item {\tt dropout\_1} $\in [0, 0.8]$, linear scale
    \item {\tt dropout\_2} $\in [0, 0.8]$, linear scale
    \item {\tt dropout\_3} $\in [0, 0.8]$, linear scale
    \item {\tt learning\_rate} $\in [0.0001, 0.01]$, log scale
    \item {\tt beta\_1} $\in [0.7, 0.99]$, log scale
    \item {\tt beta\_2} $\in [0.9, 0.999]$, log scale
    \item {\tt batch\_size} $\in [16, 512]$, log scale
\end{itemize}
\textbf{Network architecture:} 
\begin{verbatim}
        x_in = Input(shape=(32, 32, 3))

        x = Conv2D(filters=64, kernel_size=2, padding='same', use_bias=False)(x_in)
        x = BatchNormalization()(x)
        x = Activation("relu")(x)
        x = MaxPooling2D(pool_size=2)(x)
        x = Dropout(dropout_1)(x)

        x = Conv2D(filters=32, kernel_size=2, padding='same', use_bias=False)(x)
        x = BatchNormalization()(x)
        x = Activation("relu")(x)
        x = MaxPooling2D(pool_size=2)(x)
        x = Dropout(dropout_2)(x)

        x = Flatten()(x)
        x = Dense(256, activation='relu')(x)
        x = Dropout(dropout_3)(x)
        x_out = Dense(2, activation='softmax')(x)

        cnn = Model(inputs=x_in, outputs=x_out)

        adam = optimizers.Adam(learning_rate=learning_rate, 
                               beta_1=beta_1, 
                               beta_2=beta_2)
        cnn.compile(loss='categorical_crossentropy', 
                    optimizer=adam,
                    metrics=['accuracy'])
\end{verbatim}

%% file: sm_secs/4-parallel.tex
\section{Batch Construction by Sequential Penalization}
For global optimization, parallelizing BO algorithms to 
sugggest a batch of $q$ inputs, $\{x\n1,...,x^{n+q}\}$,
has been approached in multiple ways. 
For acquisitions functions that compute an expectation over future outcomes
$\P[y\n1|x\n1]$, (EI, KG, ES, MES), the acquisition value of a batch
can be computed using the expectation over multiple correlated outcomes
$\P[y\n1, ...,y^{n+q}|x\n1,...,x^{n+q}]$. This larger $q$ dimensional
expectation, effectively looking $q$ steps into the future, must be estimated 
by Monte-Carlo. At the same time, it is a function of all $q$ points
in the batch and must be optimized simultaneously over $q$ times more
dimensions $X^q$. This method of parallelization
quickly becomes infeasible for even moderate dimensions and batch sizes.
As before, adapting the method to
conditional optimisation adds another layer of Monte-Carlo
integration over $s\in S$ multiplying the computational cost.

\begin{figure*}[ht]
    \centering
    \includegraphics[width=0.8\textwidth]{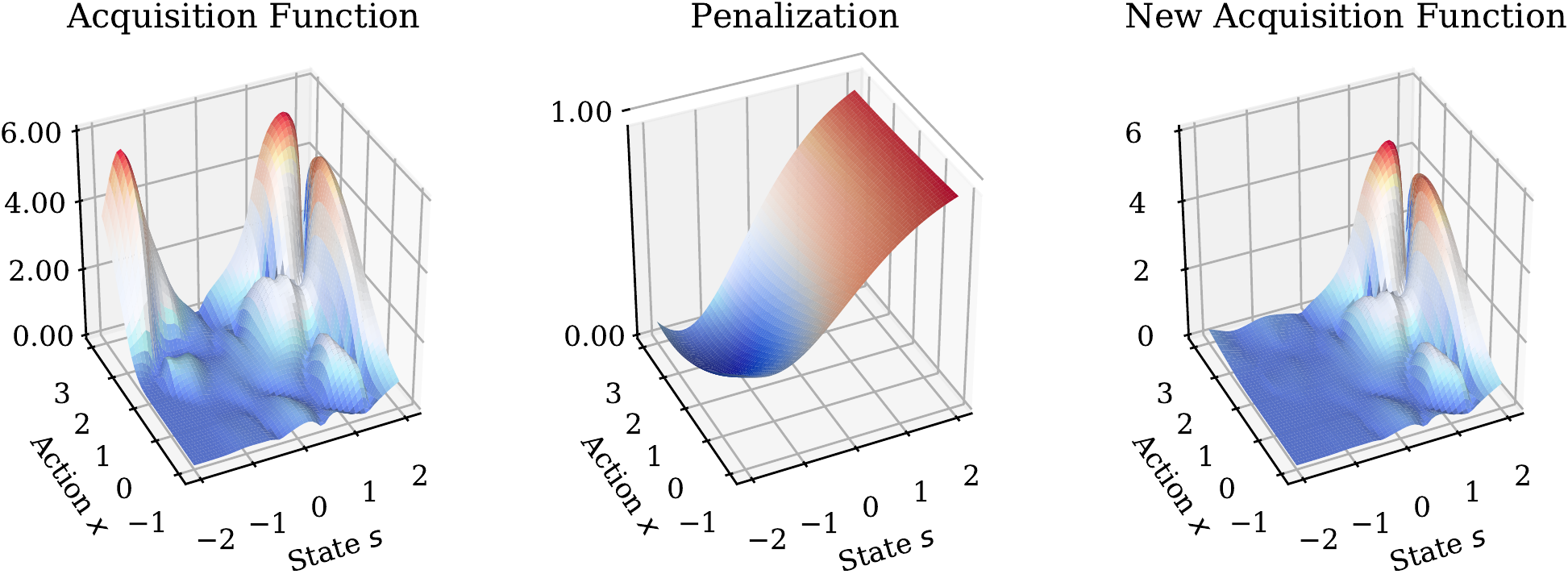}
    \caption{
    Left: acquisition function over $S\times X$ with a peak at
    $(s,x)\n1=(-1.5, 2.5)$, the first point in the batch.
    Centre: the penalization function that down-weighs any point
    $(s,x)\in S\times X$ according to similarity with $(s,x)\n1$.
    Right: The product of acquisition and penalization functions, the
    peak at $(s,x)^{n+2} = (1.6, 3.0)$ is the second point in the batch.
    }
    \label{fig:penalization}
\end{figure*}

Thompson sampling (TS) randomly suggests the next point to 
evaluate, $x\n1$. TS also has the convenient mathematical
property that $q$ step look ahead is equivalent to generating 
$q$ i.i.d samples~\citep{hernandez2017parallel, kandasamy2018parallelised}.
This property was used by \cite{char2019offline} to parallelize MTS.
\\
Alternatively, sequential construction of a batch can be done in $O(q)$ time and
we consider the method of \cite{gonzalez2016batch}. First $x\n1$ is found by
optimizing the chosen acquisition function $x\n1 = \argmax\alpha(x)$. The acquisition function
is then multiplied by a non-negative
penalty function $\phi(x, x\n1)$ that penalizes $x$ similar to $x\n1$. The next point is 
found by $x^{n+2} = \argmax_x \alpha(x)\phi(x\n1, x)$, then 
$x^{n+3} = \argmax_x \alpha(x)\phi(x\n1, x)\phi(x^{n+2}, x)$ until a batch of $q$ points is 
constructed. We use the inverted GP kernel as the penalty function
\begin{equation*}
    \phi((s,x), (s,x)^i) = 1 - \frac{k^0((s,x), (s,x)^i)}{k^0((s,x)^i,(s,x)^i)}.
\end{equation*}
See Figure~\ref{fig:penalization} for an illustration.
Previous work~\citep{groves2018parallelizing} showed
that the conditional optimization setting is well suited to
this construction method. Two points on dissimilar
states do not interact and if they are both local peaks of 
the chosen acquisition function $\alpha(s, x)$,
then both may be evaluated in parallel.
In conditional optimization, the presence of
state variables introduces multiple objective functions allowing a batch of points
to be more spread out reducing interactions and possible inefficiencies. This can be
achieved by penalization thus sidestepping the need
for expensive nested Monte-Carlo integration.
By contrast, in global optimization all $q$ points are ``crammed'' into
a single state, all interacting with the same objective requiring more care in 
batch construction techniques.

This method for batch construction may be applied to any acquisition function,
in our experiments we apply it to REVI and ConBO.

In practice, we optimize the acquisition function using multi start gradient ascent and
keep the entire history of evaluations $\{(s_t,x_t,\alpha(s_t,x_t))\}_{t=1}^{\#calls}$. Since this history
is very likely to contain multiple peaks, we simply apply the penalization to the set of
past evaluations avoiding the need to re-optimize the penalized acquisition function.
Therefore, efficiently parallelizing a conditional BO algorithm can be done in just a few
additional lines of code.

\begin{figure*}[h]
    \centering
    \includegraphics[width=0.73\textwidth]{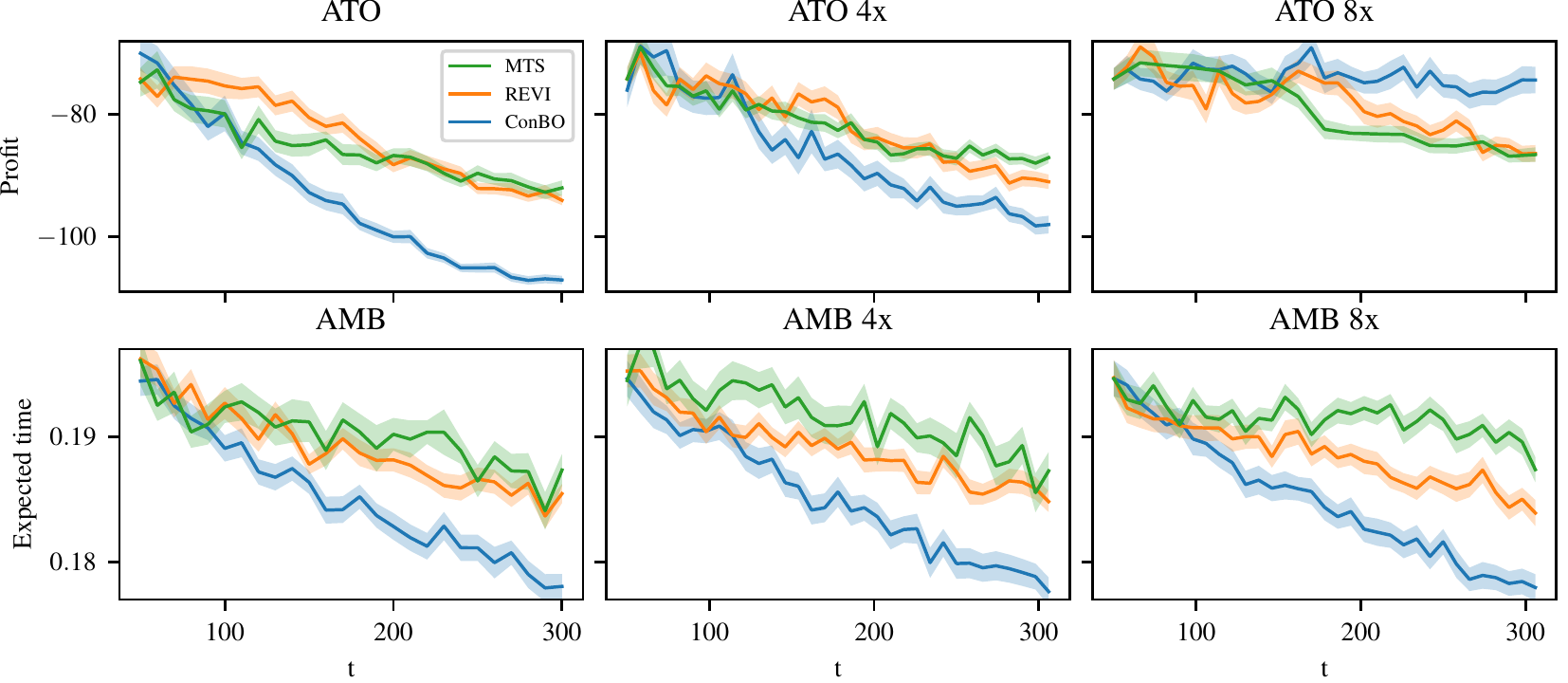}
    \caption{Algorithm performance on the Assemble To Order (top) and Ambulances (bottom) benchmarks. Serial, parallel 4 and 8.
    }
    \label{fig:real_parallel}
\end{figure*}

%% file: sm_secs/0-GLOBAL-OPTIM.tex
\section{Global Optimization}
\begin{figure}[htbp]
    \centering
    \includegraphics[width=\textwidth]{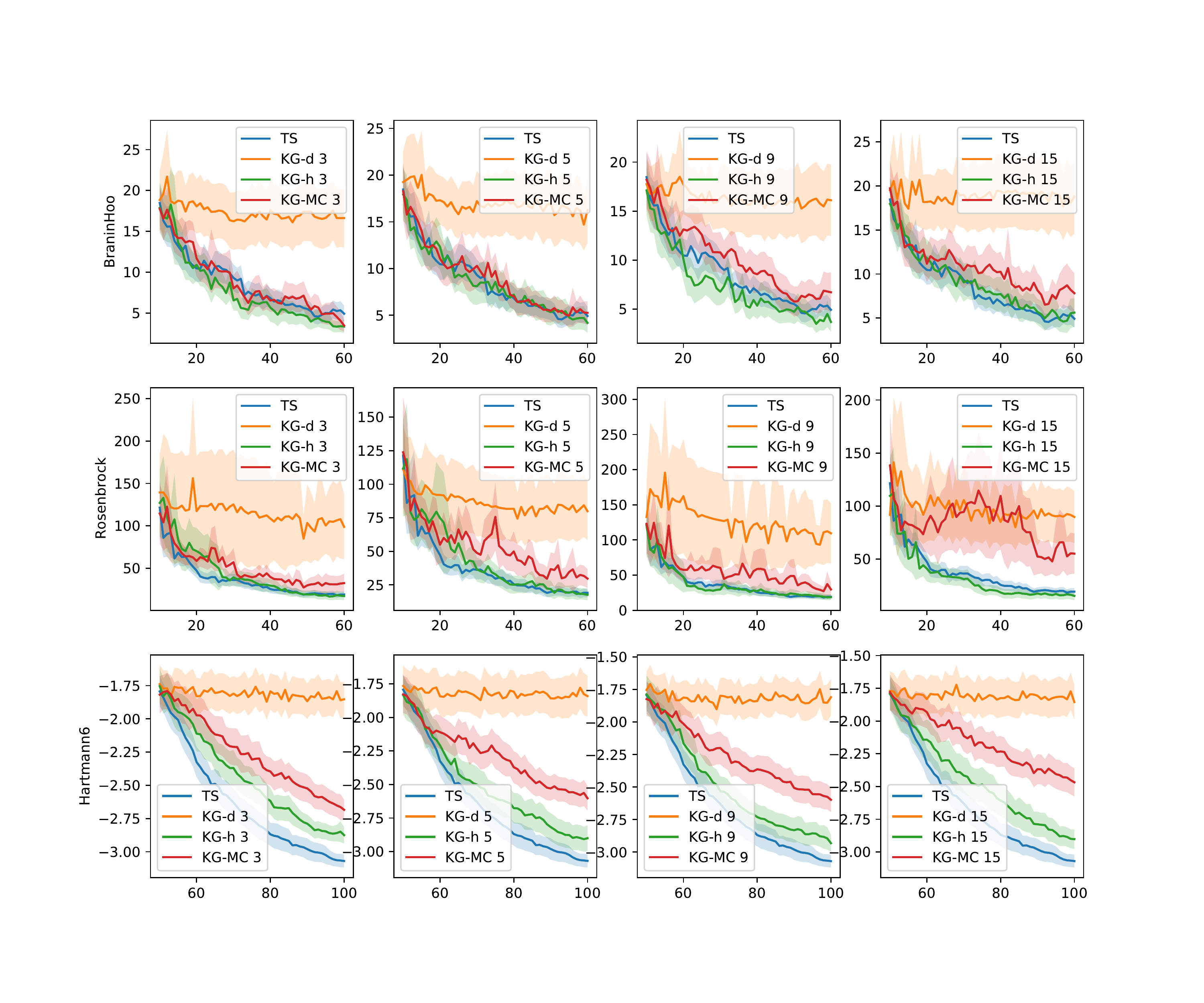}
    \caption{The KG by discretization, $\KG_d$, uses $n_z$ random points in the search space $X$ and performs
    worst by far. $\KG_{MC}$ uses $n_z$ samples of the future posterior mean function and performs well on small
    problems but suffers from unreliable noisy convergence. $\KG_h$ consistently outperforms other KG methods 
    and is the most reliable on all problems matching Thompson sampling but still performing worse on the Hartmann6
    function. This is in contrast to the conditional setting with multiple states wheer integrated KG methods outperform 
    Thomson sampling methods.
    }
    \label{fig:GO_KG}
\end{figure}

We perform a parameter sweep over the $n_z$ for each KG implementation. As a baseline we consider Thompson Sampling.
For test functions we take the popular Branin-Hoo, Rosenbrock and Hartmann6. See Figure~\ref{fig:GO_KG} for results.



%% file: sm_secs/1-ES_MES.tex
\section{Entropy Based Methods for Conditional Bayesian Optimization}
Given a GP model and a dataset, $\P[\xs|\tXn, Y^n]$ is the distribution over the peak of realizations
of GP sample functions (abusing notation) $\P[\xs|\tXn, Y^n]=\text{argmax}_x GP(\mu^n(x), k^n(x,x'))$.
Given a new sample input $x\n1$, the outcome $\P[y\n1|x\n1, \tXn, Y^n]$
is also a random variable that is Gaussian.
For thsi section, to reduce cluttering notation, we suppress 
the dependence on $\tXn, Y^n$. The mutual information between random
variables $y\n1$ and $x^*$ is defined as
\begin{equation}
    \MI(x) = \int_{x^*}\int_{y\n1} \log\left( \frac{\P[y\n1, x^*]}{\P[y\n1]\P[x^*]} \right)\P[y\n1, x^*] dy\n1 dx^*
\end{equation}
where $\P[y\n1]$ depends upon $x\n1$ yet we drop it for convenience.

\subsection{Entropy Search}
The Entropy search algorithm decomposes the above expression using $\P[y\n1, x^*]=\P[y\n1]\P[x^*|y\n1]$
resulting in the following acquisition function
\begin{eqnarray}
    \ES(x) 
    &=& \int_{\xs} \log\left(\P[\xs]\right) \P[\xs] d\xs + \int_\yn1 \int_{\xs} \log\left(\P[\xs\yn1]\right) \P[\xs|\yn1] d\xs \P[\yn1]d\yn1 \\
     &=& H[\xs] - \int_\yn1 H[\xs|y\n1] \P[\yn1]d\yn1 
\end{eqnarray}
where $H[\xs]$ is the entropy of the distribution $\P[\xs]$. For the conditional case,
the outcome $\P[y\n1|(s,x)\n1]$ is still a Gaussian random variable, and we measure the mutual information
with the peak $\xsi$ \emph{over actions constrained to a given state} $\{s_i\}\times X$ that is 
$\P[\xsi] = \text{argmax}_x GP(\mu^n(s_i,x), k^n(s_i,x,s_i,x'))$. And the conditional
entropy search acquisition function is simply
\begin{equation}
    \ES_c(s_i; (s,x)\n1) = H[\xsi] - \int_\yn1 H[\xsi| y\n1] \P[\yn1]d\yn1.
\end{equation}

\subsection{Predictive Entropy Search}
We again drop the dependence on $x\n1$ in $\P[y\n1|x\n1]$. 
The Predictive Entropy search algorithm uses an alternative decomposition of
the Mutual Information using $\P[y\n1, x^*]=\P[y\n1|\xs]\P[x^*]$
resulting in the following acquisition function
\begin{eqnarray}
    \PES(x) &=& H[\yn1] - \int_\xs H[y\n1|\xs] \P[\xs]d\xs.
\end{eqnarray}
For the conditional case,
the outcome $\P[y\n1|(s,x)\n1]$ is still a Gaussian random variable, and we measure
the mutual information with the peak $\xs$ \emph{constrained to a given state} $s_i$
that is as above 
$\P[\xsi] = \text{argmax}_x GP(\mu^n(s_i,x), k^n(s_i,x,s_i,x'))$. And the conditional
predictive entropy search acquisition function is simply
\begin{equation}
    \PES_c(s_i; (s,x)\n1) = H[\yn1] - \int_\xsi H[y\n1|\xsi] \P[\xsi]d\xs
\end{equation}
where the expression $H[y\n1|\xsi]$ is the (non Gaussian) distribution of $y\n1$ at $(s,x)\n1$ given that
the peak of state $s_i$ is at $\xsi$.

\subsection{Max-value Entropy Search}
Max-value entropy search instead measures the mutual information between the new outcome $\P[y\n1|x\n1]$
and the largest possible outcome (again abusing notation) $\P[y^*] = \max_x GP(\mu^n(x), k^n(x,x'))$, the peak value of 
posterior sample functions. The acquisition function decomposes the mutual information into
\begin{equation}
    \MES(x) = H[y\n1] - \int_{y^*} H[y\n1|y^*] dy^*
\end{equation}
The conditional version measures the mutual information between $\P[y\n1|(s,x)\n1]$ and 
the largest $y$ value amongst all outcomes with the same state
$\P[\ysi] = \max_x GP(\mu^n(s_i,x), k^n(s_i,x,s_i,x'))$
\begin{equation}
    \MES(s_i, (s,x)\n1) = H[y\n1] - \int_{\ysi} H[y\n1|\ysi] \P[\ysi]d\ysi
\end{equation}

%% file: main.bbl
\begin{thebibliography}{39}
\providecommand{\natexlab}[1]{#1}
\providecommand{\url}[1]{\texttt{#1}}
\expandafter\ifx\csname urlstyle\endcsname\relax
  \providecommand{\doi}[1]{doi: #1}\else
  \providecommand{\doi}{doi: \begingroup \urlstyle{rm}\Url}\fi

\bibitem[Bardenet et~al.(2013)Bardenet, Brendel, K{\'e}gl, and
  Sebag]{bardenet2013collaborative}
R.~Bardenet, M{\'a}ty{\'a}s Brendel, Bal{\'a}zs K{\'e}gl, and Michele Sebag.
\newblock Collaborative hyperparameter tuning.
\newblock In \emph{International Conference on Machine Learning}, pages
  199--207, 2013.

\bibitem[Char et~al.(2019)Char, Chung, Neiswanger, Kandasamy, Nelson, Boyer,
  Kolemen, and Schneider]{char2019offline}
Ian Char, Youngseog Chung, Willie Neiswanger, Kirthevasan Kandasamy,
  Andrew~Oakleigh Nelson, Mark Boyer, Egemen Kolemen, and Jeff Schneider.
\newblock Offline contextual bayesian optimization.
\newblock In \emph{Advances in Neural Information Processing Systems}, pages
  4629--4640, 2019.

\bibitem[Chung et~al.(2020)Chung, Char, Neiswanger, Kandasamy, Nelson, Boyer,
  Kolemen, and Schneider]{chung2020offline}
Youngseog Chung, Ian Char, Willie Neiswanger, Kirthevasan Kandasamy,
  Andrew~Oakleigh Nelson, Mark~D Boyer, Egemen Kolemen, and Jeff Schneider.
\newblock Offline contextual bayesian optimization for nuclear fusion.
\newblock \emph{arXiv preprint arXiv:2001.01793}, 2020.

\bibitem[Dong et~al.(2017)Dong, Eckman, Zhao, Henderson, and
  Poloczek]{dong2017empirically}
Naijia~Anna Dong, David~J Eckman, Xueqi Zhao, Shane~G Henderson, and Matthias
  Poloczek.
\newblock Empirically comparing the finite-time performance of
  simulation-optimization algorithms.
\newblock In \emph{2017 Winter Simulation Conference (WSC)}, pages 2206--2217.
  IEEE, 2017.

\bibitem[Feurer et~al.(2015)Feurer, Springenberg, and
  Hutter]{feurer2015initializing}
Matthias Feurer, Jost~Tobias Springenberg, and Frank Hutter.
\newblock Initializing bayesian hyperparameter optimization via meta-learning.
\newblock In \emph{Twenty-Ninth AAAI Conference on Artificial Intelligence},
  2015.

\bibitem[Frazier et~al.(2009)Frazier, Powell, and Dayanik]{Frazier2009}
P.~Frazier, W.~Powell, and S.~Dayanik.
\newblock The knowledge-gradient policy for correlated normal beliefs.
\newblock \emph{INFORMS Journal on Computing}, 21\penalty0 (4):\penalty0
  599--613, 2009.
\newblock ISSN 10919856.

\bibitem[Ginsbourger et~al.(2014{\natexlab{a}})Ginsbourger, Baccou, Chevalier,
  Perales, Garland, and Monerie]{Ginsbourger2014}
D.~Ginsbourger, J.~Baccou, C.~Chevalier, F.~Perales, N.~Garland, and Y~Monerie.
\newblock Bayesian adaptive reconstruction of profile optima and optimizers.
\newblock \emph{{SIAM/ASA} Journal on Uncertainty Quantification}, 2\penalty0
  (1):\penalty0 490--510, 2014{\natexlab{a}}.

\bibitem[Ginsbourger et~al.(2014{\natexlab{b}})Ginsbourger, Baccou, Chevalier,
  Perales, Garland, and Monerie]{ginsbourger2014bayesian}
David Ginsbourger, Jean Baccou, Cl{\'e}ment Chevalier, Fr{\'e}d{\'e}ric
  Perales, Nicolas Garland, and Yann Monerie.
\newblock Bayesian adaptive reconstruction of profile optima and optimizers.
\newblock \emph{SIAM/ASA Journal on Uncertainty Quantification}, 2\penalty0
  (1):\penalty0 490--510, 2014{\natexlab{b}}.

\bibitem[Gonz{\'a}lez et~al.(2016)Gonz{\'a}lez, Dai, Hennig, and
  Lawrence]{gonzalez2016batch}
Javier Gonz{\'a}lez, Zhenwen Dai, Philipp Hennig, and Neil Lawrence.
\newblock Batch bayesian optimization via local penalization.
\newblock In \emph{Artificial intelligence and statistics}, pages 648--657,
  2016.

\bibitem[Groves et~al.(2018)Groves, Pearce, and
  Branke]{groves2018parallelizing}
Matthew Groves, Michael Pearce, and Juergen Branke.
\newblock On parallelizing multi-task bayesian optimization.
\newblock In \emph{2018 Winter Simulation Conference (WSC)}, pages 1993--2002.
  IEEE, 2018.

\bibitem[Hern{\'a}ndez-Lobato et~al.(2017)Hern{\'a}ndez-Lobato, Requeima,
  Pyzer-Knapp, and Aspuru-Guzik]{hernandez2017parallel}
Jos{\'e}~Miguel Hern{\'a}ndez-Lobato, James Requeima, Edward~O Pyzer-Knapp, and
  Al{\'a}n Aspuru-Guzik.
\newblock Parallel and distributed thompson sampling for large-scale
  accelerated exploration of chemical space.
\newblock In \emph{Proceedings of the 34th International Conference on Machine
  Learning-Volume 70}, pages 1470--1479. JMLR. org, 2017.

\bibitem[Huang et~al.(2006{\natexlab{a}})Huang, Allen, Notz, and
  Miller]{huang2006sequentialMF}
Deng Huang, Theodore~T Allen, William~I Notz, and R~Allen Miller.
\newblock Sequential kriging optimization using multiple-fidelity evaluations.
\newblock \emph{Structural and Multidisciplinary Optimization}, 32\penalty0
  (5):\penalty0 369--382, 2006{\natexlab{a}}.

\bibitem[Huang et~al.(2006{\natexlab{b}})Huang, Allen, Notz, and
  Miller]{huang2006sequential}
Deng Huang, TT~Allen, WI~Notz, and RA~Miller.
\newblock Sequential kriging optimization using multiple-fidelity evaluations.
\newblock \emph{Structural and Multidisciplinary Optimization}, 32\penalty0
  (5):\penalty0 369--382, 2006{\natexlab{b}}.

\bibitem[Jeong et~al.(2005)Jeong, Murayama, and Yamamoto]{jeong2005efficient}
Shinkyu Jeong, Mitsuhiro Murayama, and Kazuomi Yamamoto.
\newblock Efficient optimization design method using kriging model.
\newblock \emph{Journal of aircraft}, 42\penalty0 (2):\penalty0 413--420, 2005.

\bibitem[Jones et~al.(1998)Jones, Schonlau, and Welch]{Jones1998}
D.~R. Jones, M.~Schonlau, and W.~J. Welch.
\newblock Efficient global optimization of expensive black-box functions.
\newblock \emph{Journal of Global optimization}, 13\penalty0 (4):\penalty0
  455--492, 1998.
\newblock ISSN 09255001.

\bibitem[Kandasamy et~al.(2015)Kandasamy, Schneider, and
  P{\'o}czos]{kandasamy2015high}
Kirthevasan Kandasamy, Jeff Schneider, and Barnab{\'a}s P{\'o}czos.
\newblock High dimensional bayesian optimisation and bandits via additive
  models.
\newblock In \emph{International Conference on Machine Learning}, pages
  295--304, 2015.

\bibitem[Kandasamy et~al.(2016)Kandasamy, Dasarathy, Oliva, Schneider, and
  P{\'o}czos]{kandasamy2016gaussian}
Kirthevasan Kandasamy, Gautam Dasarathy, Junier~B Oliva, Jeff Schneider, and
  Barnab{\'a}s P{\'o}czos.
\newblock Gaussian process bandit optimisation with multi-fidelity evaluations.
\newblock In \emph{Advances in Neural Information Processing Systems}, pages
  992--1000, 2016.

\bibitem[Kandasamy et~al.(2017)Kandasamy, Dasarathy, Schneider, and
  P{\'o}czos]{kandasamy2017multi}
Kirthevasan Kandasamy, Gautam Dasarathy, Jeff Schneider, and Barnab{\'a}s
  P{\'o}czos.
\newblock Multi-fidelity bayesian optimisation with continuous approximations.
\newblock In \emph{Proceedings of the 34th International Conference on Machine
  Learning-Volume 70}, pages 1799--1808. JMLR. org, 2017.

\bibitem[Kandasamy et~al.(2018)Kandasamy, Krishnamurthy, Schneider, and
  P{\'o}czos]{kandasamy2018parallelised}
Kirthevasan Kandasamy, Akshay Krishnamurthy, Jeff Schneider, and Barnab{\'a}s
  P{\'o}czos.
\newblock Parallelised bayesian optimisation via thompson sampling.
\newblock In \emph{International Conference on Artificial Intelligence and
  Statistics}, pages 133--142, 2018.

\bibitem[Kingma and Ba(2014)]{kingma2014adam}
Diederik~P Kingma and Jimmy Ba.
\newblock Adam: A method for stochastic optimization.
\newblock \emph{arXiv preprint arXiv:1412.6980}, 2014.

\bibitem[Krause and Ong(2011)]{krause2011contextual}
Andreas Krause and Cheng~S Ong.
\newblock Contextual gaussian process bandit optimization.
\newblock In \emph{Advances in neural information processing systems}, pages
  2447--2455, 2011.

\bibitem[Morales-Enciso and Branke(2015)]{MoralesBranke15}
S.~Morales-Enciso and J.~Branke.
\newblock Tracking global optima in dynamic environments with efficient global
  optimization.
\newblock \emph{European Journal of Operational Research}, 242:\penalty0
  744--755, 2015.

\bibitem[Paul et~al.(2018)Paul, Osborne, and Whiteson]{paul2018contextual}
Supratik Paul, Michael~A Osborne, and Shimon Whiteson.
\newblock Contextual policy optimisation.
\newblock \emph{arXiv preprint arXiv:1805.10662}, 2018.

\bibitem[Pearce and Branke(2016)]{PearceBranke1}
M.~Pearce and J.~Branke.
\newblock Efficient information collection on portfolios.
\newblock Technical report, University of Warwick, 2016.

\bibitem[Pearce and Branke(2018)]{pearce2018continuous}
Michael Pearce and Juergen Branke.
\newblock Continuous multi-task bayesian optimisation with correlation.
\newblock \emph{European Journal of Operational Research}, 270\penalty0
  (3):\penalty0 1074--1085, 2018.

\bibitem[Pearce et~al.(2019)Pearce, Poloczek, and Branke]{pearce2019bayesian}
Michael Pearce, Matthias Poloczek, and Juergen Branke.
\newblock Bayesian optimization allowing for common random numbers.
\newblock \emph{arXiv preprint arXiv:1910.09259}, 2019.

\bibitem[Perrone et~al.(2018)Perrone, Jenatton, Seeger, and
  Archambeau]{perrone2018scalable}
Valerio Perrone, Rodolphe Jenatton, Matthias~W Seeger, and C{\'e}dric
  Archambeau.
\newblock Scalable hyperparameter transfer learning.
\newblock In \emph{Advances in Neural Information Processing Systems}, pages
  6845--6855, 2018.

\bibitem[Picheny and Ginsbourger(2013)]{picheny2013nonstationary}
Victor Picheny and David Ginsbourger.
\newblock A nonstationary space-time gaussian process model for partially
  converged simulations.
\newblock \emph{SIAM/ASA Journal on Uncertainty Quantification}, 1\penalty0
  (1):\penalty0 57--78, 2013.

\bibitem[Poloczek et~al.(2016)Poloczek, Wang, and Frazier]{poloczek2016warm}
Matthias Poloczek, Jialei Wang, and Peter~I Frazier.
\newblock Warm starting bayesian optimization.
\newblock In \emph{2016 Winter Simulation Conference (WSC)}, pages 770--781.
  IEEE, 2016.

\bibitem[Poloczek et~al.(2017)Poloczek, Wang, and Frazier]{poloczek2017multi}
Matthias Poloczek, Jialei Wang, and Peter Frazier.
\newblock Multi-information source optimization.
\newblock In \emph{Advances in Neural Information Processing Systems}, pages
  4289--4299, 2017.

\bibitem[Rasmussen and Williams(2004)]{rasmussen2006gaussian}
C.~E. Rasmussen and C.~K.~I. Williams.
\newblock \emph{Gaussian Processes for Machine Learning}.
\newblock MIT Press, 2004.
\newblock ISBN 026218253X.

\bibitem[Sambakh{\'e} et~al.(2019)Sambakh{\'e}, Rouan, Bacro, and
  Goz{\'e}]{sambakhe2019conditional}
Diari{\'e}tou Sambakh{\'e}, Lauriane Rouan, Jean-No{\"e}l Bacro, and Eric
  Goz{\'e}.
\newblock Conditional optimization of a noisy function using a kriging
  metamodel.
\newblock \emph{Journal of Global Optimization}, 73\penalty0 (3):\penalty0
  615--636, 2019.

\bibitem[Scott et~al.(2011)Scott, Frazier, and Powell]{scott2011correlated}
Warren Scott, Peter Frazier, and Warren Powell.
\newblock The correlated knowledge gradient for simulation optimization of
  continuous parameters using gaussian process regression.
\newblock \emph{SIAM Journal on Optimization}, 21\penalty0 (3):\penalty0
  996--1026, 2011.

\bibitem[Smith-Miles et~al.(2014)Smith-Miles, Baatar, Wreford, and
  Lewis]{Smith-Miles2014}
Kate Smith-Miles, Davaatseren Baatar, Brendan Wreford, and Rhyd Lewis.
\newblock {Towards objective measures of algorithm performance across instance
  space}.
\newblock \emph{Computers and Operations Research}, 45:\penalty0 12--24, 2014.
\newblock ISSN 03050548.
\newblock \doi{10.1016/j.cor.2013.11.015}.
\newblock URL \url{http://dx.doi.org/10.1016/j.cor.2013.11.015}.

\bibitem[Snoek et~al.(2012)Snoek, Larochelle, and Adams]{snoek2012practical}
Jasper Snoek, Hugo Larochelle, and Ryan~P Adams.
\newblock Practical bayesian optimization of machine learning algorithms.
\newblock In \emph{Advances in neural information processing systems}, pages
  2951--2959, 2012.

\bibitem[Toscano-Palmerin and Frazier(2018)]{toscano2018bayesian}
Saul Toscano-Palmerin and Peter~I Frazier.
\newblock Bayesian optimization with expensive integrands.
\newblock \emph{arXiv preprint arXiv:1803.08661}, 2018.

\bibitem[Wang and Jegelka(2017)]{wang2017max}
Zi~Wang and Stefanie Jegelka.
\newblock Max-value entropy search for efficient bayesian optimization.
\newblock In \emph{Proceedings of the 34th International Conference on Machine
  Learning-Volume 70}, pages 3627--3635. JMLR. org, 2017.

\bibitem[Wu and Frazier(2017)]{wu2017discretization}
Jian Wu and Peter~I Frazier.
\newblock Discretization-free knowledge gradient methods for bayesian
  optimization.
\newblock \emph{arXiv preprint arXiv:1707.06541}, 2017.

\bibitem[Xie et~al.(2016)Xie, Frazier, and Chick]{xie2016bayesian}
Jing Xie, Peter~I Frazier, and Stephen~E Chick.
\newblock Bayesian optimization via simulation with pairwise sampling and
  correlated prior beliefs.
\newblock \emph{Operations Research}, 64\penalty0 (2):\penalty0 542--559, 2016.

\end{thebibliography}
